\documentclass{article}
\usepackage{iclr2026_conference,times}
\iclrfinalcopy

\usepackage{amsmath,amsfonts,bm}

\def\eqref#1{equation~\ref{#1}}

\def\1{\bm{1}}

\DeclareMathAlphabet{\mathsfit}{\encodingdefault}{\sfdefault}{m}{sl}
\SetMathAlphabet{\mathsfit}{bold}{\encodingdefault}{\sfdefault}{bx}{n}

\usepackage{latexsym}
\usepackage{amsmath}
\usepackage{pifont}
\usepackage{amsfonts}
\usepackage{colortbl}
\usepackage{tcolorbox}
\usepackage{amssymb}
\usepackage{graphicx}
\usepackage{booktabs}
\usepackage{array}
\usepackage{multirow}
\usepackage{tabularx}
\usepackage{listings}
\usepackage{courier}
\usepackage[table]{xcolor}
\usepackage{hyperref}
\usepackage{pifont}
\usepackage{subcaption}
\usepackage{wrapfig}
\usepackage{enumitem}

\newcommand{\cmark}{\textcolor{green!60!black}{\ding{51}}}
\newcommand{\xmark}{\textcolor{red}{\ding{55}}}
\newcommand{\ours}{JigShape}

\title{JigShape: Evaluating Visual-Geometric Reasoning in VLMs through Jigsaw Puzzles}

\author{
Shawn Li\textsuperscript{1},
Wei Yang\textsuperscript{1},
Jike Zhong\textsuperscript{1},
Jiate Li\textsuperscript{1},
Jiawei Yang\textsuperscript{1},
You Qin\textsuperscript{2}, 
Ryan Rossi\textsuperscript{3},\\
\textbf{
Franck Dernoncourt\textsuperscript{3},
Roger Zimmermann\textsuperscript{2},
Yue Wang\textsuperscript{1},
Zhengzhong Tu\textsuperscript{4},
} \\
\textbf{
Vicente Ordóñez\textsuperscript{5},
Mohit Bansal\textsuperscript{6},
Yue Zhao\textsuperscript{1}}
\\
\textsuperscript{1}University of Southern California,
\textsuperscript{2}National University of Singapore, \\
\textsuperscript{3}Adobe Research, 
\textsuperscript{4}Texas A\&M University,
\textsuperscript{5}Rice University, \\
\textsuperscript{6}The University of North Carolina at Chapel Hill
}

\begin{document}
\maketitle

\begin{abstract}
Jigsaw puzzle solving requires jointly reasoning about visual content and geometric constraints, yet existing benchmarks use rectangular cuts that create ambiguous ground truth in texture-repeated regions. We introduce \textit{\ours{}}, a benchmark with tab-and-blank interlocking pieces where geometric constraints provide strong local compatibility requirements that, combined with visual content, yield unambiguous ground truth. Across 95K instances at four grid densities (4$\times$4 to 16$\times$16), we find that \textbf{zero-shot VLMs largely lack geometric reasoning}: only one of five frontier models (GPT-5.5) exceeds random baseline on 4$\times$4 puzzles, while all others perform at chance level. While supervised fine-tuning achieves $>$97\% on 4$\times$4, \textbf{all models collapse on larger grids}: GPT-5.5 drops from 70\% to near-random on 8$\times$8, and even fine-tuned models fall below 5\% on 12$\times$12. This ``scaling cliff'' suggests current architectures cannot maintain consistent constraint satisfaction as the number of pieces increases. \ours{} establishes scalable geometric reasoning as an open challenge for vision-language models.
\end{abstract}

\centerline{\small \textbf{Dataset:} \url{https://huggingface.co/datasets/ShawnLi02/JigShape-Train}}

\section{Introduction}

Vision-language models (VLMs) have advanced rapidly in recent years, achieving strong performance on recognition, captioning, visual question answering, and multimodal reasoning~\citep{openai2023gpt4v,anthropic2024claude,team2023gemini,liu2023llava,li2023blip2,dai2023instructblip}. These capabilities have enabled applications ranging from document understanding to embodied agents. A natural question arises: \emph{Can VLMs perform spatial reasoning?} Recent studies suggest this remains challenging: VLMs struggle with basic spatial relations~\citep{liu2023vsr,kamath2023whatsup}, depth perception~\citep{fu2024blink}, and 3D understanding~\citep{chen2024spatialvlm}. Spatial reasoning, the ability to infer geometric relationships and reconstruct coherent structures from fragmented observations, is fundamental to many real-world tasks including robotics manipulation, architectural design, and scientific visualization.

\begin{figure}[t]
\centering
\includegraphics[width=0.98\linewidth]{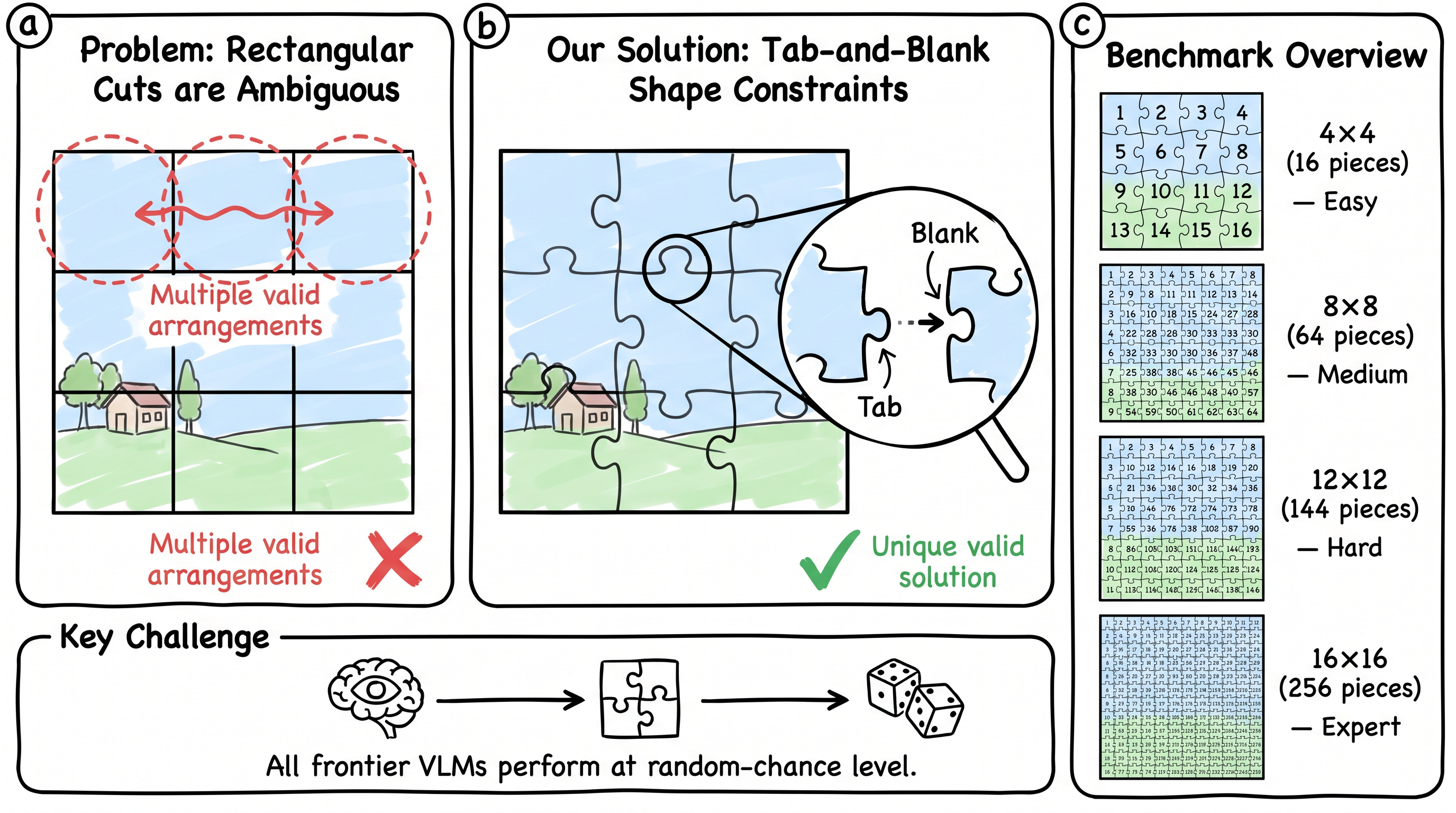}
\vspace{-0.1in}
\caption{Overview of \ours{}. \textbf{Left}: Rectangular cuts create ambiguous ground truth in repeated-texture regions; our tab-and-blank constraints enforce local compatibility, yielding unambiguous solutions when combined with visual content. \textbf{Right}: Benchmark taxonomy with four grid densities (4$\times$4 to 16$\times$16) and shape vs.\ no-shape ablation for controlled evaluation.}
\label{fig:teaser}
\vspace{-0.3cm}
\end{figure}

Jigsaw puzzles provide a natural testbed for spatial reasoning. Solving a puzzle requires integrating local visual cues (texture, color, object boundaries) with global structural constraints (pieces must tile the image plane without gaps or overlaps). Unlike semantic tasks where partial understanding suffices, jigsaw solving demands precise spatial localization: each piece has exactly one correct position. This makes puzzles an ideal probe for measuring whether VLMs can reason about geometric arrangements rather than merely recognizing visual patterns. Notably, prior work in self-supervised learning has shown that training models to solve jigsaw puzzles yields visual representations that transfer well to downstream tasks such as object detection and image classification~\citep{noroozi2016unsupervised,carlucci2019domain,chen2023jigsawvit,ren2023masked}. This suggests that jigsaw-solving ability is not merely a narrow skill but reflects broader visual reasoning capabilities.

\noindent \textbf{Current Benchmarks.}
Recent work has begun evaluating VLMs on jigsaw-style tasks. These benchmarks divide images into rectangular patches on a grid (e.g., 2$\times$2 or 3$\times$3), shuffle the patches, and ask models to predict the original arrangement. Jigsaw-Puzzles~\citep{lyu2025jigsawpuzzles} designs five cognitive tasks around 2$\times$2 and 3$\times$3 grids; the best model (Gemini-2.5-Pro) reaches 77\% overall but drops to 30\% on open-ended order generation, exposing a gap between multiple-choice recognition and unconstrained spatial reconstruction. The JPwLEG dataset~\citep{song2023sd2rl} provides 3$\times$3 and 5$\times$5 puzzles with eroded gaps, on which specialized solvers such as VLHSA~\citep{xu2025vlhsa} and PuzLM~\citep{elkin2025puzlm} are trained and evaluated. Visual Jigsaw~\citep{wu2025visualjigsaw} uses jigsaw solving as a post-training task to improve MLLM visual understanding across images, videos, and 3D data. These efforts establish spatial reasoning as a challenging frontier for VLMs.

However, existing benchmarks share a common design: they all use simple rectangular cuts. This leads to two fundamental limitations that compromise their diagnostic value (Tab.~\ref{tab:benchmark_comparison}).

\begin{itemize}[leftmargin=*]
    \item \textbf{Ambiguity from rectangular cuts.} When images are divided into rectangular patches, regions with uniform texture (sky, grass, walls) become visually indistinguishable. Consider a 3$\times$3 puzzle of a landscape: the three sky patches may be interchangeable without any perceptual difference. Multiple permutations are thus visually valid, yet ground-truth labels accept only one. This makes evaluation ill-posed: a ``wrong'' prediction may be perceptually reasonable, conflating genuine reasoning failures with label arbitrariness. Benchmarks using rectangular cuts cannot distinguish whether models fail due to lack of spatial reasoning or due to ambiguous supervision.
    \item \textbf{Low grid density.} Existing benchmarks operate at coarse granularity, typically 2$\times$2 to 5$\times$5 patches~\citep{lyu2025jigsawpuzzles,song2023sd2rl}. A 3$\times$3 grid has only 9! $\approx$ 363K permutations, while an 8$\times$8 grid has 64! $\approx 10^{89}$. Coarse grids limit the difficulty ceiling and may be solvable through simple heuristics (e.g., matching edge colors). Dense grids are necessary to stress-test spatial reasoning at scale and to measure how performance degrades with combinatorial complexity.
\end{itemize}

We introduce \textit{\ours{}}, a large-scale benchmark that addresses both limitations. \ours{} features physically interlocking jigsaw pieces with \emph{tab-and-blank geometric constraints}, as illustrated in Fig.~\ref{fig:teaser}. Each piece boundary is either a convex tab or a concave blank; adjacent pieces must have complementary edges (tab fits into blank). This design, inspired by real jigsaw puzzles, provides strong local compatibility constraints that substantially reduce the valid search space. Combined with distinct visual content in each piece, these constraints yield unambiguous ground truth, converting jigsaw evaluation from an ill-posed to a well-posed problem and enabling rigorous measurement of joint visual-geometric reasoning. We verify that every puzzle instance has a unique solution using a classical constraint-based solver (Appendix~\ref{app:classical_solver}).

The benchmark spans four grid densities (4$\times$4, 8$\times$8, 12$\times$12, 16$\times$16), with 95,468 total instances generated from 23,742 high-resolution source images (Tab.~\ref{tab:grid_stats}). To isolate the effect of geometric constraints, we additionally generate a \emph{no-shape} (rectangular) version of the test set. Comparing performance with and without shape constraints directly measures whether VLMs can utilize boundary geometry, addressing a central question: \emph{Can VLMs jointly reason over visual content and geometric constraints?}

Our evaluation reveals three key findings. First, \textbf{zero-shot geometric reasoning is largely absent}: among five frontier models (six configurations), only GPT-5.5 exceeds random baseline on 4$\times$4 puzzles (69.65\% vs.\ 6.25\%), while others, including reasoning-enabled variants like Claude Opus 4.8 and Grok-4.2, fail entirely. Supervised fine-tuning on Qwen3-VL-8B~\citep{bai2025qwen25vl} and Gemma3-12B~\citep{team2024gemma} achieves $>$97\% on 4$\times$4, demonstrating the task is learnable. Second, \textbf{all models collapse on larger grids}: GPT-5.5 drops from 70\% to 4\% on 8$\times$8, and even fine-tuned models fall to $<$5\% on 12$\times$12. This ``scaling cliff'' suggests current architectures cannot maintain consistent constraint satisfaction as the number of pieces increases. Third, \textbf{SFT models rely primarily on geometric constraints}: removing tab-blank boundaries causes Qwen3-VL-8B to drop from 97\% to 10\% PA, barely above random, indicating models learn to exploit shape cues with limited visual content integration.

Our contributions:
\begin{itemize}
    \item \textbf{Problem formalization}: We cast jigsaw puzzle solving as \emph{joint visual-geometric reasoning}, where tab-and-blank boundaries provide local compatibility constraints that, combined with visual content, yield unambiguous ground truth and convert an ill-posed visual matching task into a well-posed reasoning problem.
    \item \textbf{Large-scale benchmark}: \ours{} provides 95K+ instances across four grid densities (4$\times$4 to 16$\times$16), with a shape vs.\ no-shape ablation that isolates the contribution of geometric constraints.
    \item \textbf{Comprehensive evaluation}: We evaluate five frontier VLMs in six zero-shot configurations and two fine-tuned models, revealing that (1) only GPT-5.5 exceeds random baseline on 4$\times$4, (2) fine-tuning achieves near-perfect 4$\times$4 accuracy, proving the task is learnable, and (3) all models collapse on 8$\times$8 and beyond.
\end{itemize}

\section{Related Work}

\noindent \textbf{Spatial Reasoning in VLMs.}
Spatial reasoning has emerged as a key challenge for vision-language models. VSR~\citep{liu2023vsr} evaluates binary spatial relations (``is A left of B?'') and finds that even large VLMs struggle with basic positional reasoning. SpatialBench~\citep{chen2024spatialvlm} extends this to 3D spatial understanding, revealing significant gaps in depth and distance estimation. BLINK~\citep{fu2024blink} tests relative depth ordering and shows that models often fail on perceptually obvious comparisons. What's Up~\citep{kamath2023whatsup} systematically probes spatial relation understanding, finding that VLMs exhibit consistent biases and fail to generalize across viewpoints. Other benchmarks probe compositional reasoning~\citep{thrush2022winoground,li2026fortisbenchmarkingoverprivilegeagent,li2026defensespromptattackslearn,parcalabescu2022valse} and visual question answering~\citep{johnson2017clevr,Li_2025_CVPR,Li_Ji_Wu_Li_Qin_Wei_Zimmermann_2024,li2025secureondevicevideoood,hudson2019gqa}, revealing broader gaps in structured visual understanding. These benchmarks evaluate \emph{recognition} of existing spatial relations. In contrast, \ours{} tests \emph{reconstruction}: assembling a coherent structure from fragments by satisfying geometric constraints, a fundamentally harder task requiring integration of local cues with global consistency.

\noindent \textbf{Jigsaw Puzzle Benchmarks for VLMs.}
Recent work has adopted jigsaw puzzles as a testbed for spatial reasoning in VLMs. Jigsaw-Puzzles~\citep{lyu2025jigsawpuzzles} designs five cognitive tasks around 2$\times$2 and 3$\times$3 grids on 1,100 real-world images; the best model reaches 77\% overall accuracy but only 30\% on open-ended order generation. VGRP-Bench~\citep{ren2025vgrp} evaluates 20 rule-based constraint puzzles (e.g., Sudoku, Kakuro, Nonogram) on a grid, testing logical reasoning rather than spatial reconstruction. LEGO-Puzzles~\citep{tang2025lego} tests sequential assembly of block structures, requiring step-by-step spatial planning. Visual Jigsaw~\citep{wu2025visualjigsaw} proposes using jigsaw puzzle solving as a self-supervised post-training task to enhance MLLM visual perception; while they demonstrate that training on jigsaw tasks improves downstream performance, their work focuses on training rather than evaluation and uses rectangular cuts without geometric constraints. All existing benchmarks use rectangular cuts, which create ambiguous ground truth when images contain repeated textures (sky, grass, uniform surfaces). Multiple arrangements may be perceptually equivalent, yet only one is marked correct, conflating reasoning failures with label arbitrariness. \ours{} addresses this fundamental limitation through geometric shape constraints that ensure unique solutions.

\noindent \textbf{Self-Supervised Representation Learning.}
Jigsaw puzzle solving was introduced as a self-supervised pretext task by \citet{noroozi2016unsupervised} and later shown to improve domain generalization~\citep{carlucci2019domain} and Vision Transformer robustness~\citep{chen2023jigsawvit,li-etal-2025-treble,li2026geometrydensityfewshotcrossdomain,ren2023masked}. More recently, jigsaw puzzles have emerged as a training signal for large vision-language models: Visual Jigsaw~\citep{wu2025visualjigsaw} uses jigsaw solving as RL post-training, improving fine-grained perception and 3D understanding; Jigsaw-R1~\citep{wang2025jigsawr1} finds that rule-based visual RL on jigsaw generalizes more effectively than supervised fine-tuning; PuzzleCraft~\citep{jeddi2025puzzlecraft} introduces exploration-aware curriculum learning with puzzle-based RLVR; and Spatial-SSRL~\citep{liu2025spatialssrl} derives self-supervised spatial signals: including shuffled patch reordering, to improve spatial understanding via GRPO. These works demonstrate that jigsaw training improves general model capabilities beyond puzzle solving, motivating benchmarks like \ours{} that serve both as evaluation instruments and as training data sources for spatial reasoning.

\noindent \textbf{Computational Jigsaw Puzzle Solving.}
The computational puzzle-solving literature spans decades. Classical approaches use pairwise compatibility measures based on color gradients and texture continuity~\citep{pomeranz2011fully}, genetic algorithms for large puzzles~\citep{sholomon2013genetic}, and loop constraints for global consistency~\citep{son2014solving}. Recent deep learning methods include SD2RL~\citep{song2023sd2rl}, which introduces the JPwLEG dataset and a Siamese DQN for puzzles with eroded gaps, VLHSA~\citep{xu2025vlhsa}, which applies vision-language alignment to the same setting, PuzLM~\citep{elkin2025puzlm}, which formulates puzzle solving as sequence-to-sequence prediction, and DiffAssemble~\citep{scarpellini2024diffassemble}, which uses graph diffusion for 2D and 3D reassembly. These methods train specialized models on puzzle data (typically 3$\times$3 to 5$\times$5 grids). Our work differs in two ways: (1) we evaluate \emph{zero-shot} reasoning in general-purpose VLMs rather than training puzzle-specific models, and (2) we introduce geometric shape constraints as a benchmark dimension, enabling controlled ablation of visual vs.\ geometric reasoning.

\begin{table*}[t]
\centering
\caption{Comparison with existing jigsaw-style benchmarks. Prior work uses rectangular cuts with limited grid sizes and lacks geometric constraints. \ours{} introduces tab-and-blank shape constraints that, combined with visual content, ensure unambiguous ground truth, plus dense grids up to 16$\times$16 and controlled shape vs.\ no-shape ablation.}
\label{tab:benchmark_comparison}
\resizebox{\textwidth}{!}{
\begin{tabular}{lccccccc}
\toprule
\textbf{Benchmark} &
\begin{tabular}{@{}c@{}} \textbf{Task} \\ \textbf{Type} \end{tabular} &
\begin{tabular}{@{}c@{}} \textbf{Grid} \\ \textbf{Sizes} \end{tabular} &
\begin{tabular}{@{}c@{}} \textbf{Max} \\ \textbf{Pieces} \end{tabular} &
\begin{tabular}{@{}c@{}} \textbf{Shape} \\ \textbf{Constraint} \end{tabular} &
\begin{tabular}{@{}c@{}} \textbf{Guaranteed} \\ \textbf{Unique} \end{tabular} &
\begin{tabular}{@{}c@{}} \textbf{Shape} \\ \textbf{Ablation} \end{tabular} &
\textbf{Scale} \\
\midrule
Jigsaw-Puzzles~\citep{lyu2025jigsawpuzzles} & VLM Eval & 2$\times$2, 3$\times$3 & 9 & \xmark & \xmark & \xmark & 1.1K \\
JPwLEG~\citep{song2023sd2rl} & Model Training & 3$\times$3, 5$\times$5 & 25 & \xmark & \xmark & \xmark & 12K \\
Visual Jigsaw~\citep{wu2025visualjigsaw} & Model Training & 3$\times$3 & 9 & \xmark & \xmark & \xmark & 118K \\
Jigsaw-R1~\citep{wang2025jigsawr1} & Model Training & 2$\times$1 to 2$\times$2 & 4 & \xmark & \xmark & \xmark & 83K \\
\midrule
\ours{} (Ours) & VLM Eval & 4$\times$4 to 16$\times$16 & \textbf{256} & \cmark & \cmark & \cmark & \textbf{95K} \\
\bottomrule
\end{tabular}
}
\vspace{-0.3cm}
\end{table*}

\section{Benchmark}

\ours{} is a geometry-centric benchmark where adjacent pieces must have complementary edge shapes (tab matches blank). These geometric constraints, combined with distinct visual content in each piece, yield unambiguous ground truth. This section describes the formal task definition, shape constraint mechanism, and dataset construction.

\subsection{Task Formulation}

Given a shuffled set of jigsaw pieces with visible shape boundaries, the task is to predict the correct grid position for each piece. Formally, a puzzle instance is a tuple $P = (I, G, E, \pi)$, where $I$ is the source image, $G$ is the $N \times N$ grid, $E: \mathcal{E} \rightarrow \{\text{tab}, \text{blank}, \text{flat}\}$ assigns edge types to piece boundaries, and $\pi: \{0, \ldots, N^2-1\} \rightarrow \{0, \ldots, N^2-1\}$ is a random permutation mapping piece IDs to grid positions.

The model receives an image showing all $N^2$ pieces arranged in a grid layout ordered by piece ID (not by correct position), with each piece clearly labeled with its ID number. The model must output a mapping from each piece ID to its predicted $(row, col)$ position in the solved puzzle.

The \emph{geometric compatibility constraint} requires that adjacent pieces have complementary touching edges:
\begin{equation}
    E(p_i, \text{right}) = \text{tab} \Leftrightarrow E(p_j, \text{left}) = \text{blank}.
\end{equation}
Shape constraints drastically reduce the valid search space by enforcing local compatibility. Combined with distinct visual content in each piece, these constraints yield a unique ground truth. Unlike rectangular-cut puzzles where repeated textures create multiple valid-looking arrangements, our tab-and-blank design ensures every instance has exactly one correct solution.

\subsection{Shape Constraint Mechanism}
\label{sec:shape}

Real jigsaw puzzles use interlocking shapes to ensure pieces fit together unambiguously. We implement this mechanism with three edge types:

\begin{itemize}[leftmargin=*]
    \item \textbf{Tab}: A convex semicircular protrusion extending outward from the piece boundary. Tabs carry visual content from the neighboring piece, creating a natural ``lock'' that must fit into a corresponding blank.
    \item \textbf{Blank}: A concave semicircular indentation cut into the piece boundary. Blanks receive the tab from an adjacent piece, forming the complementary half of the interlock.
    \item \textbf{Flat}: A straight edge appearing only on the outer boundary of the puzzle. Corner pieces have two flat edges; edge pieces have one; interior pieces have none.
\end{itemize}

Each piece's \emph{edge signature} $(e_{\text{top}}, e_{\text{right}}, e_{\text{bottom}}, e_{\text{left}})$ encodes its geometric constraints. For an $N \times N$ puzzle, there are $(N-1) \times N$ horizontal internal edges and $N \times (N-1)$ vertical internal edges, each randomly assigned tab-blank polarity during generation. This randomization ensures that edge signatures vary across instances, preventing models from memorizing positional patterns.

The shape constraints provide two types of information for solving: (1) \emph{local compatibility}, where adjacent pieces must have complementary edges, and (2) \emph{global position cues}, where flat edges indicate boundary positions and the number of flat edges distinguishes corners (2), edges (1), and interior pieces (0).


\subsection{Grid Configurations}

We construct puzzles at four density levels to measure how spatial reasoning scales with complexity (Tab.~\ref{tab:grid_stats}). The grid sizes are chosen to span a wide range of combinatorial difficulty while remaining visually parseable:

\begin{itemize}[leftmargin=*]
    \item \textbf{4$\times$4 (16 pieces)}: Entry-level difficulty with $\sim 2 \times 10^{13}$ permutations. Pieces are large enough that visual content provides strong cues. This setting tests basic spatial reasoning.
    \item \textbf{8$\times$8 (64 pieces)}: Medium difficulty with $\sim 10^{89}$ permutations. Individual pieces contain less distinctive content, increasing reliance on shape constraints and local edge matching.
    \item \textbf{12$\times$12 (144 pieces)}: High difficulty with $\sim 10^{249}$ permutations. Pieces are small and often visually similar, making geometric constraints essential for disambiguation.
    \item \textbf{16$\times$16 (256 pieces)}: Extreme difficulty with $\sim 10^{507}$ permutations. This configuration stress-tests the limits of current VLM architectures.
\end{itemize}

Tab geometry is scaled with grid density: the tab ratio (tab depth as a fraction of piece side length) increases from 0.15 at 4$\times$4 to 0.28 at 16$\times$16, ensuring tabs remain visually perceptible as piece size decreases. The final layout sizes range from 724$\times$744 pixels (4$\times$4) to 3292$\times$3312 pixels (16$\times$16), all within the input resolution limits of modern VLMs.

\begin{table}[t]
\centering
\caption{Grid configurations in \ours{}. Tab ratio indicates tab depth as a fraction of piece side length. Permutations show the combinatorial search space.}
\label{tab:grid_stats}
\begin{tabular}{lccccc}
\toprule
\textbf{Grid} & \textbf{Pieces} & \textbf{Tab Ratio} & \textbf{Layout Size} & \textbf{Permutations} & \textbf{Difficulty} \\
\midrule
4$\times$4 & 16 & 0.15 & 724$\times$744 & $\sim 2 \times 10^{13}$ & Easy \\
8$\times$8 & 64 & 0.20 & 1532$\times$1552 & $\sim 10^{89}$ & Medium \\
12$\times$12 & 144 & 0.24 & 2412$\times$2432 & $\sim 10^{249}$ & Hard \\
16$\times$16 & 256 & 0.28 & 3292$\times$3312 & $\sim 10^{507}$ & Expert \\
\bottomrule
\end{tabular}
\vspace{-0.3cm}
\end{table}

\begin{figure}[t]
\centering
\begin{tabular}{cc}
\includegraphics[width=0.45\linewidth]{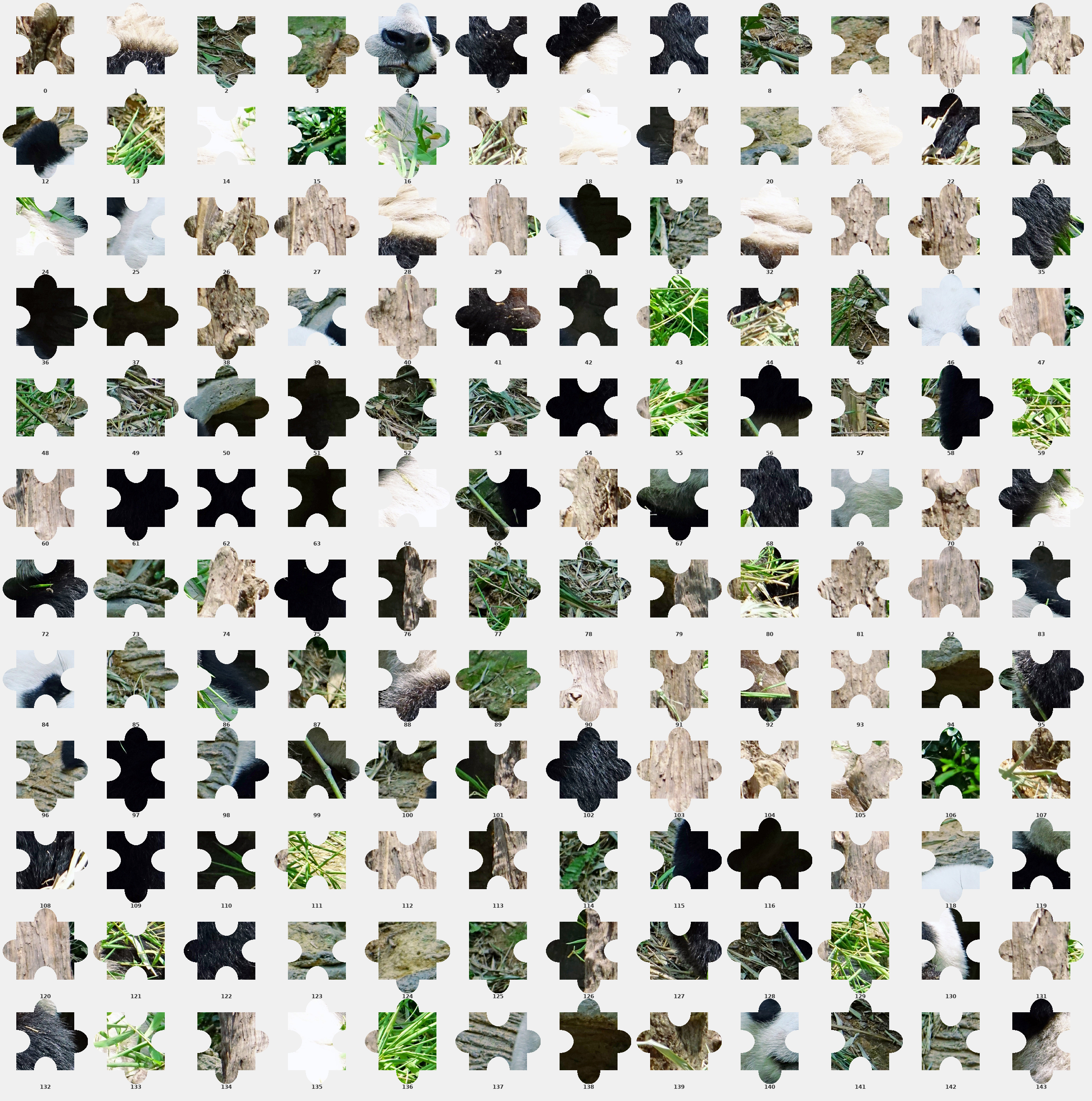} &
\includegraphics[width=0.45\linewidth]{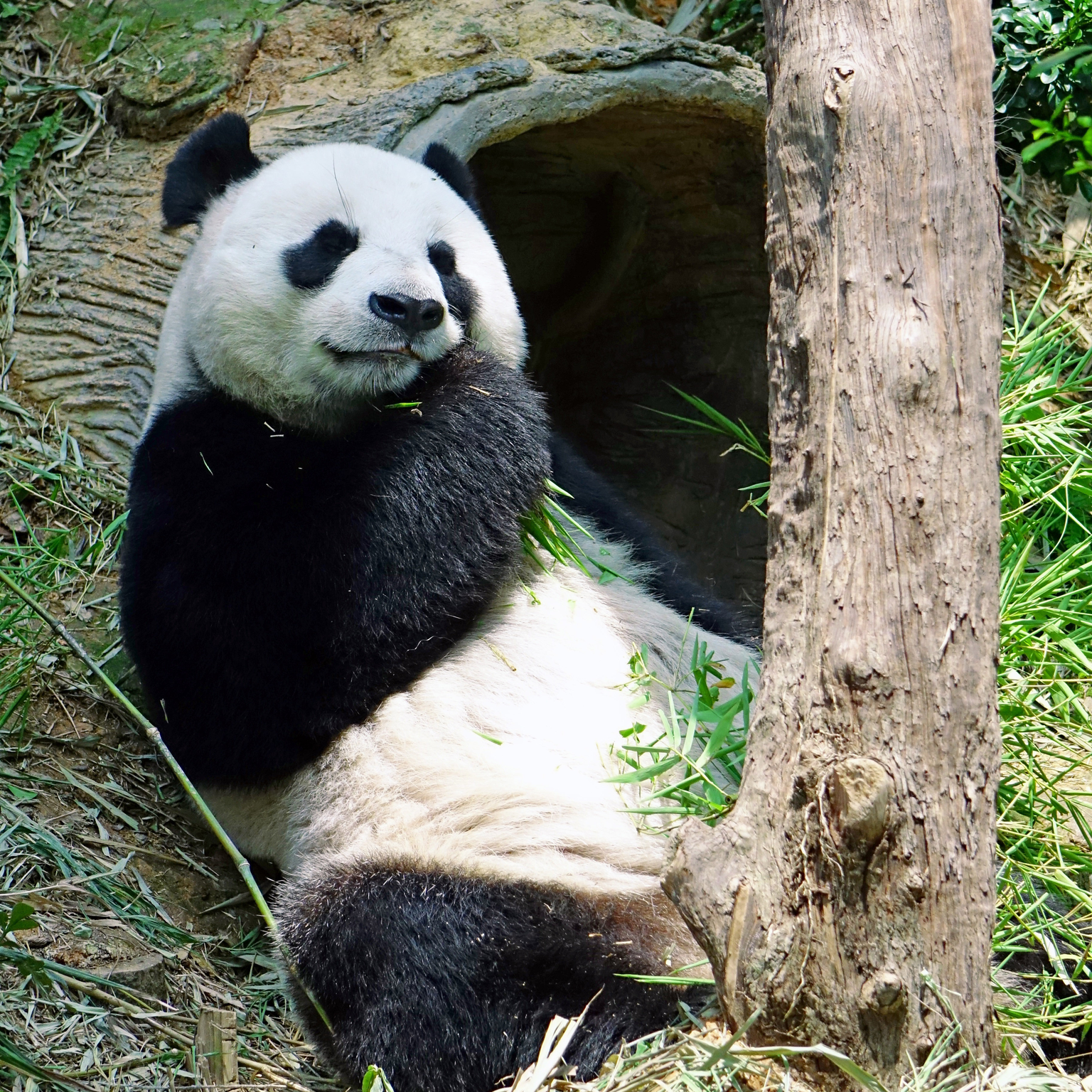} \\
\end{tabular}
\caption{Example 12$\times$12 puzzle instance from \ours{}. Left: shuffled layout with 144 labeled pieces showing tab-and-blank edge shapes. Right: original source image representing ground truth. Geometric constraints (tabs must match blanks) enforce local compatibility, while visual content disambiguates pieces with identical edge signatures. See Appendix~\ref{app:examples} for examples at other grid sizes.}
\label{fig:example_main}
\end{figure}

\subsection{Source Images and Statistics}

High-resolution source images are essential for dense grid configurations, as each piece must contain sufficient visual detail for potential matching. We curate images from three sources:

\begin{itemize}[leftmargin=*]
    \item \textbf{DIV2K}~\citep{agustsson2017div2k}: 900 images at $\sim$2K resolution, originally designed for super-resolution research. These images feature diverse natural scenes with rich textures.
    \item \textbf{DIV8K}~\citep{gu2019div8k}: 1,500 images at up to 8K resolution, providing extremely high detail suitable for our densest 16$\times$16 configuration.
    \item \textbf{Unsplash}~\citep{unsplash}: 21,342 curated high-resolution photographs covering landscapes, architecture, objects, and abstract patterns. This forms the majority of our dataset and ensures broad visual diversity.
\end{itemize}

In total, 23,742 unique source images are used. Each image generates one puzzle instance per grid configuration, yielding 95,468 total instances. We split the data by source image to prevent train-test leakage: the same image never appears in both training and evaluation sets, even at different grid sizes (Tab.~\ref{tab:data_stats}). All experimental results reported in this paper are on the \textbf{evaluation set}; the held-out \textbf{test set} is reserved for a future public competition to enable fair comparison of methods developed after publication.

\noindent \textbf{No-shape version for ablation.}
We create a \textbf{no-shape} ablation set of 500 paired instances at 4$\times$4 resolution. Each pair shares the same source image and piece ID permutation, differing only in edge type: the shape version uses tab-and-blank edges, while the no-shape version uses rectangular cuts. This paired design enables controlled measurement of whether models utilize geometric boundary information. We restrict this ablation to 4$\times$4 because larger grids with rectangular cuts lack unique ground truth: regions with repeated textures (sky, grass, walls) become interchangeable, making multiple arrangements visually valid. See Appendix~\ref{app:pipeline} for generation details.

\begin{table}[t]
\centering
\caption{Benchmark split statistics. Each source image generates one instance per grid size. For ablation, a no-shape (rectangular) version is provided only for 4$\times$4, as larger grids without shape constraints have ambiguous ground truth. \textbf{The held-out test set is reserved for a future competition.}}
\label{tab:data_stats}
\begin{tabular}{lrrrrr}
\toprule
\textbf{Split} & \textbf{4$\times$4} & \textbf{8$\times$8} & \textbf{12$\times$12} & \textbf{16$\times$16} & \textbf{Total} \\
\midrule
Train & 22,992 & 22,992 & 22,992 & 22,992 & 91,968 \\
Eval & 250 & 250 & 250 & 250 & 1,000 \\
Test (held-out) & 500 & 500 & 500 & 500 & 2,000 \\
Test (No-Shape) & 500 & --- & --- & --- & 500 \\
\midrule
\textbf{Total} & 24,242 & 23,742 & 23,742 & 23,742 & \textbf{95,468} \\
\bottomrule
\end{tabular}
\vspace{-0.3cm}
\end{table}

\subsection{Generation Pipeline}

\paragraph{Image Preprocessing.}
The source image is resized so both dimensions are divisible by the grid size $N$, yielding piece dimensions $h \times w$. We use high-quality Lanczos resampling~\citep{lanczos} to preserve detail.

\paragraph{Edge Assignment.}
For each internal edge between adjacent grid cells, we randomly assign tab-blank polarity with equal probability, while boundary edges are set to flat. This random assignment ensures diverse edge configurations across instances, preventing models from memorizing fixed patterns.

\paragraph{Shape Mask and Content Extraction.}
We generate an alpha mask for each piece encoding its shape. Tabs are semicircular protrusions with radius $r = \text{tab\_ratio} \times \min(h, w) / 2$, centered on the edge midpoint; blanks are corresponding indentations. Piece content is extracted using these masks. Crucially, tab regions carry visual content from the neighboring cell, mimicking real jigsaw puzzles where the protruding tab shows part of the adjacent image region. This creates natural visual continuity cues that a reasoning model could exploit.

\paragraph{Layout Composition.}
Piece IDs (0 to $N^2-1$) are randomly permuted to prevent any correlation between ID and grid position. Pieces are arranged in a regular grid layout ordered by ID (not by correct position), with each piece clearly labeled with its numeric ID. Importantly, pieces are presented in their original orientation without rotation; this ensures a unique ground-truth solution exists, as allowing rotations would introduce ambiguity when pieces have symmetric shapes. The pipeline is fully deterministic given a random seed, enabling reproducibility. We verify correctness by confirming that pieces can be recomposed to reconstruct the original image exactly (Fig.~\ref{fig:example_main}; additional examples in Appendix~\ref{app:examples}).

\section{Evaluation Protocol}

\subsection{Metrics}
\label{sec:metrics}

We report four metrics (formal definitions in Appendix~\ref{app:metrics}):
\textbf{Piece Accuracy (PA)}: fraction of pieces placed in correct positions.
\textbf{Exact Match (EM)}: fraction of instances with all pieces correct (PA = 100\%).
\textbf{Adjacency Accuracy (AA)}: fraction of ground-truth adjacent piece pairs that are also adjacent in the prediction, measuring local coherence.
\textbf{Shape Compatibility (SC)}: fraction of predicted adjacencies with geometrically compatible edges (tab matches blank). The random baseline is grid-specific due to flat boundary edges (e.g., 31\% for 4$\times$4, 44\% for 16$\times$16; see Appendix~\ref{app:metrics} for exact derivation); SC above this baseline indicates shape constraint utilization.

\section{Experiments}

\paragraph{Models.}
We evaluate five frontier VLMs in six zero-shot configurations: \textbf{GPT-5.5} and \textbf{GPT-5.4-mini}~\citep{openai2023gpt4v}; \textbf{Grok-4.2} in reasoning and non-reasoning modes; \textbf{Llama-4-Maverick}; and \textbf{Claude Opus 4.8}~\citep{anthropic2024claude}. Additionally, we fine-tune \textbf{Qwen3-VL-8B}~\citep{bai2025qwen25vl} and \textbf{Gemma3-12B}~\citep{team2024gemma} on mixed-grid training data (4$\times$4: 3K, 8$\times$8: 2K, 12$\times$12: 1K, 16$\times$16: 0.5K; 6.5K total) as supervised baselines.

\paragraph{Setup.}
All models are evaluated on 250 eval samples per grid with standardized prompts (Appendix~\ref{app:prompts}). We report PA, EM, AA, SC (Sec.~\ref{sec:metrics}; formal definitions in Appendix~\ref{app:metrics}). Random baseline: PA = $1/N^2$, AA = $4/[N(N{+}1)]$, SC is grid-specific due to flat boundary edges (30.8\%--44.1\%; see Appendix~\ref{app:metrics} for exact derivation). We omit 16$\times$16 evaluation for frontier VLMs: since all zero-shot models already perform at random chance on 12$\times$12 (PA $<$ 1\%), further evaluation on 16$\times$16 would yield no additional insight. See Appendix~\ref{app:experiment} for full experiment details.

\subsection{Main Results}

\begin{table}[t]
\centering
\caption{Main results: Piece Accuracy (PA) and Exact Match (EM) on \ours{}. Higher is better. All models evaluated on 250 eval samples per grid. \textbf{Since frontier VLMs already perform poorly at the 12×12 grid, we omit the 16×16 evaluation for these models.}}
\label{tab:main_results}
\small
\begin{tabular}{l|cc|cc|cc|cc}
\toprule
& \multicolumn{2}{c|}{\textbf{4$\times$4}} & \multicolumn{2}{c|}{\textbf{8$\times$8}} & \multicolumn{2}{c|}{\textbf{12$\times$12}} & \multicolumn{2}{c}{\textbf{16$\times$16}} \\
\textbf{Model} & PA & EM & PA & EM & PA & EM & PA & EM \\
\midrule
\multicolumn{9}{l}{\textit{Frontier VLMs (Zero-Shot)}} \\
GPT-5.5 & \textbf{69.65} & \textbf{26.40} & \textbf{4.37} & 0.00 & 0.67 & 0.00 & --- & --- \\
Claude Opus 4.8 & 8.50 & 0.00 & 1.38 & 0.00 & 0.38 & 0.00 & --- & --- \\
Grok-4.2-reasoning & 11.97 & 0.00 & 1.66 & 0.00 & 0.67 & 0.00 & --- & --- \\
Grok-4.2 (non-reasoning) & 6.90 & 0.00 & 1.54 & 0.00 & 0.64 & 0.00 & --- & --- \\
GPT-5.4-mini & 6.75 & 0.00 & 1.57 & 0.00 & 0.52 & 0.00 & --- & --- \\
Llama-4-Maverick & 6.62 & 0.00 & 1.57 & 0.00 & \textbf{0.69} & 0.00 & --- & --- \\
\midrule
\multicolumn{9}{l}{\textit{Supervised Fine-Tuned Models (mixed-grid training)}} \\
Qwen3-VL-8B (SFT) & 97.28 & \textbf{90.80} & 27.34 & 0.00 & 3.44 & 0.00 & \textbf{0.40} & 0.00 \\
Gemma3-12B (SFT) & \textbf{97.82} & 89.20 & \textbf{33.58} & \textbf{0.40} & \textbf{4.57} & 0.00 & 0.35 & 0.00 \\
\midrule
Random Baseline & 6.25 & 0.00 & 1.56 & 0.00 & 0.69 & 0.00 & 0.39 & 0.00 \\
\bottomrule
\end{tabular}
\end{table}

\begin{table}[t]
\centering
\caption{Auxiliary metrics: Adjacency Accuracy (AA) and Shape Compatibility (SC) on \ours{}. AA measures local coherence; SC measures geometric constraint utilization. Random baselines are grid-specific (see Appendix~\ref{app:metrics}).}
\label{tab:auxiliary_metrics}
\small
\begin{tabular}{l|cc|cc|cc|cc}
\toprule
& \multicolumn{2}{c|}{\textbf{4$\times$4}} & \multicolumn{2}{c|}{\textbf{8$\times$8}} & \multicolumn{2}{c|}{\textbf{12$\times$12}} & \multicolumn{2}{c}{\textbf{16$\times$16}} \\
\textbf{Model} & AA & SC & AA & SC & AA & SC & AA & SC \\
\midrule
\multicolumn{9}{l}{\textit{Frontier VLMs (Zero-Shot)}} \\
GPT-5.5 & \textbf{54.22} & \textbf{71.33} & \textbf{10.58} & \textbf{42.47} & 2.90 & 42.28 & --- & --- \\
Claude Opus 4.8 & 18.02 & 33.78 & 4.92 & 39.42 & 2.65 & 42.14 & --- & --- \\
Grok-4.2-reasoning & 22.22 & 37.62 & 5.50 & 39.58 & 2.66 & \textbf{42.41} & --- & --- \\
Grok-4.2 (non-reasoning) & 17.22 & 32.88 & 5.75 & 39.76 & 2.74 & 42.18 & --- & --- \\
GPT-5.4-mini & 18.48 & 33.09 & 4.92 & 39.58 & 2.58 & 42.34 & --- & --- \\
Llama-4-Maverick & 16.40 & 33.60 & 5.06 & 39.46 & \textbf{2.85} & 42.24 & --- & --- \\
\midrule
\multicolumn{9}{l}{\textit{Supervised Fine-Tuned Models (mixed-grid training)}} \\
Qwen3-VL-8B (SFT) & 95.77 & 98.40 & 15.58 & 70.75 & 0.59 & 52.11 & \textbf{0.00} & \textbf{45.02} \\
Gemma3-12B (SFT) & \textbf{97.35} & \textbf{98.95} & \textbf{38.44} & \textbf{87.06} & \textbf{10.88} & \textbf{62.10} & 0.00 & 45.12 \\
\midrule
Random Baseline & 20.00 & 30.83 & 5.56 & 38.99 & 2.56 & 42.34 & 1.47 & 44.13 \\
\bottomrule
\end{tabular}
\end{table}

\paragraph{GPT-5.5 is the only zero-shot model to exceed random baseline.}
Tab.~\ref{tab:main_results} shows that GPT-5.5 achieves 69.65\% PA and 26.40\% EM on 4$\times$4 puzzles, far exceeding random baseline (6.25\% PA). Tab.~\ref{tab:auxiliary_metrics} confirms this with high AA (54.22\% vs.\ 20.0\% random) and SC (71.33\% vs.\ 30.8\% random), indicating GPT-5.5 leverages both visual content and geometric constraints. In contrast, all other zero-shot models, including Claude Opus 4.8, Grok-4.2-reasoning, and Llama-4-Maverick, achieve near-random PA (6.62--11.97\%) with 0\% EM. Their 4$\times$4 SC values (32.9--37.6\%) are close to the random baseline (30.8\%), indicating they do not meaningfully utilize shape constraints.

\paragraph{Extended reasoning provides limited benefit.}
Grok-4.2-reasoning outperforms its non-reasoning variant by 5 percentage points on 4$\times$4 (11.97\% vs.\ 6.90\% PA), but this remains far below GPT-5.5 and insufficient for reliable puzzle solving. The improvement vanishes on larger grids. This suggests that current reasoning mechanisms help process visual-spatial information but do not enable geometric constraint satisfaction.

\paragraph{All zero-shot models fail on larger grids.}
GPT-5.5's advantage disappears at higher complexity: 4.37\% PA on 8$\times$8 and 0.67\% PA on 12$\times$12, both near random baseline. All zero-shot models converge to chance-level performance as grid density increases. On 8$\times$8 and 12$\times$12, SC values (39.4--42.5\%) match the grid-specific random baselines (39.0\% and 42.3\%, respectively), confirming that models place pieces without regard for edge compatibility on larger grids.

\paragraph{SFT models learn the task but exhibit a scaling cliff.}
Both SFT models achieve $>$97\% PA and $\sim$90\% EM on 4$\times$4, confirming the task is learnable with supervision. Gemma3-12B outperforms Qwen3-VL-8B across all metrics, particularly on 8$\times$8 (33.58\% vs.\ 27.34\% PA; 38.44\% vs.\ 15.58\% AA). However, both exhibit sharp performance degradation: 8$\times$8 drops to 27--34\% PA, 12$\times$12 to 3--5\% PA, and 16$\times$16 to near-random (0.35--0.40\% PA vs.\ 0.39\% baseline).

\subsection{Shape vs.\ No-Shape Ablation}

To directly measure whether VLMs utilize geometric constraints, we compare performance on identical puzzles with and without tab-blank boundaries. The no-shape ablation uses rectangular pieces on the same source images with identical piece ID permutations. We conduct this ablation only on 4$\times$4 grids because larger grids with rectangular cuts have ambiguous ground truth (multiple visually-valid arrangements exist).

\begin{table}[t]
\centering
\caption{Shape vs.\ No-Shape ablation on 4$\times$4 puzzles. ``Shape'' uses tab-blank constraints; ``No-Shape'' uses rectangular pieces. $\Delta$ shows the improvement from shape constraints. PA: Piece Accuracy (\%), AA: Adjacency Accuracy (\%), EM: Exact Match (\%).}
\label{tab:ablation}
\small
\begin{tabular}{l|ccc|ccc}
\toprule
& \multicolumn{3}{c|}{\textbf{Shape}} & \multicolumn{3}{c}{\textbf{No-Shape}} \\
\textbf{Model} & PA & AA & EM & PA & AA & EM \\
\midrule
\multicolumn{7}{l}{\textit{Frontier VLMs (Zero-Shot)}} \\
GPT-5.5 & 69.65 & 54.22 & 26.40 & 23.90 & 25.69 & 0.40 \\
\midrule
\multicolumn{7}{l}{\textit{Supervised Fine-Tuned Models}} \\
Qwen3-VL-8B (SFT) & 97.28 & 95.77 & 90.80 & 10.00 & 30.97 & 0.00 \\
Gemma3-12B (SFT) & 97.82 & 97.35 & 89.20 & 13.78 & 35.48 & 0.00 \\
\midrule
Random Baseline & 6.25 & 20.00 & 0.00 & 6.25 & 20.00 & 0.00 \\
\bottomrule
\end{tabular}
\end{table}

\paragraph{Shape constraints are critical for GPT-5.5.}
Tab.~\ref{tab:ablation} shows that GPT-5.5's performance drops dramatically from 69.65\% to 23.90\% PA when shape constraints are removed, a 45.75 percentage point decrease. Exact match rate also drops from 26.40\% to 0.40\%. This confirms that GPT-5.5 actively utilizes tab-and-blank geometry for spatial reasoning. Without these constraints, GPT-5.5 still outperforms random (23.90\% vs.\ 6.25\%), indicating it also leverages visual content cues, but the geometric constraints provide the majority of its reasoning signal.

\paragraph{SFT models overfit to shape constraints.}
Both SFT models show dramatic accuracy drops when shape constraints are removed: Qwen3-VL-8B falls from 97.28\% to 10.00\% PA, and Gemma3-12B from 97.82\% to 13.78\%. These residual accuracies (10--14\%) only marginally exceed random baseline (6.25\%), indicating SFT models rely predominantly on geometric cues with minimal visual content reasoning. This finding highlights a potential failure mode: SFT models may overfit to the easiest discriminative features (shape boundaries) while largely ignoring visual content.

\subsection{Analysis}

\noindent \textbf{Zero-shot geometric reasoning remains largely unsolved.}
Only one of six zero-shot configurations (GPT-5.5) exceeds random baseline on 4$\times$4 puzzles, achieving 69.65\% PA with 26.40\% EM. The remaining models, including reasoning-enabled variants like Claude Opus 4.8 and Grok-4.2-reasoning, achieve SC values near the grid-specific random baseline (e.g., 32.9--37.6\% vs.\ 30.8\% random on 4$\times$4), indicating they place pieces without utilizing geometric constraints. This aligns with prior findings that VLMs struggle with spatial relations~\citep{liu2023vsr,kamath2023whatsup} and compositional reasoning~\citep{thrush2022winoground}, but reveals an even more fundamental limitation in geometric constraint satisfaction.

\noindent \textbf{The scaling cliff: from solvable to intractable.}
All models exhibit a sharp performance drop from 4$\times$4 to larger grids. GPT-5.5 falls from 69.65\% to 4.37\% PA on 8$\times$8; SFT models drop from $>$97\% to 27--34\% PA. At 12$\times$12 and 16$\times$16, all models converge to random baseline. This ``scaling cliff'' contrasts with classical puzzle solvers that scale to thousands of pieces~\citep{pomeranz2011fully,sholomon2013genetic}, suggesting that current VLM architectures, despite leveraging powerful vision encoders~\citep{dosovitskiy2021vit,radford2021clip}, cannot decompose complex puzzles into tractable subproblems or maintain consistent constraint satisfaction across many pieces.

\noindent \textbf{SFT models rely primarily on geometric constraints.}
While SFT models achieve near-perfect 4$\times$4 accuracy, the no-shape ablation reveals they depend heavily on geometric cues: Qwen3-VL-8B drops from 97.28\% to 10.00\% PA without shape constraints, while Gemma3-12B retains only 13.78\%. Both residual accuracies barely exceed random baseline (6.25\%), indicating SFT models learn to leverage shape boundaries as the primary signal, with limited integration of visual content for spatial reasoning.

\noindent \textbf{Position-based error analysis reveals flat-edge dependence.}
We decompose piece accuracy by position type: corners (2 flat edges), edges (1 flat edge), and interior pieces (0 flat edges). As shown in Tab.~\ref{tab:position_analysis}, accuracy follows a consistent pattern: corner $>$ edge $>$ interior across all grid sizes. On 4$\times$4, SFT models achieve $>$99\% on corners but only 95--96\% on interior pieces. This gap widens dramatically on larger grids: at 12$\times$12, Gemma3-12B achieves 39.70\% on corners but only 1.78\% on interior pieces. This pattern indicates models primarily rely on flat edges for positioning and struggle with interior constraint satisfaction where only tab-blank compatibility is available.

\begin{table}[t]
\centering
\caption{Position-based accuracy (\%) for SFT models. Corner pieces (2 flat edges) are easiest; interior pieces (0 flat edges) are hardest. The gap widens as grid size increases.}
\label{tab:position_analysis}
\small
\begin{tabular}{l|ccc|ccc}
\toprule
& \multicolumn{3}{c|}{\textbf{Qwen3-VL-8B}} & \multicolumn{3}{c}{\textbf{Gemma3-12B}} \\
\textbf{Grid} & Corner & Edge & Interior & Corner & Edge & Interior \\
\midrule
4$\times$4 & 99.70 & 97.00 & 95.40 & 99.40 & 97.90 & 96.10 \\
8$\times$8 & 66.60 & 37.15 & 16.43 & 78.80 & 44.30 & 21.41 \\
12$\times$12 & 34.20 & 6.28 & 1.08 & 39.70 & 8.03 & 1.78 \\
16$\times$16 & 5.10 & 0.54 & 0.26 & 1.00 & 0.29 & 0.35 \\
\bottomrule
\end{tabular}
\vspace{-0.3cm}
\end{table}

\section{Conclusion}

We introduced \textit{\ours{}}, a benchmark that formulates jigsaw puzzle solving as joint visual-geometric reasoning. Tab-and-blank shape constraints provide strong local compatibility requirements that, combined with distinct visual content, yield unambiguous ground truth and convert an ill-posed task into a well-posed evaluation. The benchmark spans four grid densities (4$\times$4 to 16$\times$16) with 95K+ instances.

Our evaluation reveals three key findings: (1) \textbf{Zero-shot geometric reasoning is largely unsolved}: only GPT-5.5 exceeds random baseline on 4$\times$4 (69.65\% PA, 26.40\% EM), while all other frontier models fail; (2) \textbf{A scaling cliff exists}: all models, including SFT models achieving $>$97\% on 4$\times$4, collapse to near-random performance on 8$\times$8 and larger grids; (3) \textbf{SFT models rely primarily on geometric constraints}: Qwen3-VL-8B drops from 97.28\% to 10.00\% PA without shape constraints, barely exceeding random baseline (6.25\%), indicating models learn to leverage geometric cues as the primary signal with limited integration of visual content.

\ours{} establishes that scalable geometric reasoning remains an open challenge. Future work should explore curriculum learning across grid sizes, training objectives that encourage joint visual-geometric integration, and architectures capable of compositional constraint satisfaction.

\section{Ethics Statement \& Reproducibility Statement}

All source images are from publicly available datasets (DIV2K, DIV8K under academic license; Unsplash under permissive license). We release the complete generation pipeline, evaluation code, and configuration to facilitate reproducibility and further research.

\bibliography{references}

@article{chen2023jigsawvit,
  author  = {Yingyi Chen and Xi Shen and Yahui Liu and Qinghua Tao and Johan A. K. Suykens},
  title   = {Jigsaw-{ViT}: Learning Jigsaw Puzzles in Vision Transformer},
  journal = {Pattern Recognition Letters},
  volume  = {166},
  pages   = {53--60},
  year    = {2023}
}

@article{wu2025visualjigsaw,
  author     = {Penghao Wu and Yushan Zhang and Haiwen Diao and Bo Li and Lewei Lu and Ziwei Liu},
  title      = {Visual Jigsaw Post-Training Improves {MLLMs}},
  journal    = {CoRR},
  volume     = {abs/2509.25190},
  year       = {2025},
  eprinttype = {arXiv},
  eprint     = {2509.25190}
}

@article{openai2023gpt4v,
  title={{GPT-4V(ision)} System Card},
  author={{OpenAI}},
  journal={OpenAI Technical Report},
  year={2023}
}

@article{anthropic2024claude,
  title={The Claude 3 Model Family: Opus, Sonnet, Haiku},
  author={{Anthropic}},
  journal={Anthropic Technical Report},
  year={2024}
}

@article{team2023gemini,
  title={Gemini: A Family of Highly Capable Multimodal Models},
  author={{Google DeepMind}},
  journal={arXiv preprint arXiv:2312.11805},
  year={2023}
}

@inproceedings{liu2023vsr,
  title={Visual Spatial Reasoning},
  author={Liu, Fangyu and Emerson, Guy and Collier, Nigel},
  booktitle={Transactions of the Association for Computational Linguistics},
  year={2023}
}

@article{chen2024spatialvlm,
  title={{SpatialVLM}: Endowing Vision-Language Models with Spatial Reasoning Capabilities},
  author={Chen, Boyuan and Xu, Zhuo and Kirmani, Sean and Ichter, Brian and Sadigh, Dorsa and Guibas, Leonidas and Xia, Fei},
  journal={arXiv preprint arXiv:2401.12168},
  year={2024}
}

@article{kamath2023whatsup,
  title={What's ``up'' with vision-language models? Investigating their struggle with spatial reasoning},
  author={Kamath, Amita and Hessel, Jack and Chang, Kai-Wei},
  journal={arXiv preprint arXiv:2310.19785},
  year={2023}
}

@article{fu2024blink,
  title={{BLINK}: Multimodal Large Language Models Can See but Not Perceive},
  author={Fu, Xingyu and Hu, Yushi and Li, Bangzheng and Feng, Yu and Wang, Haoyu and Lin, Xudong and Roth, Dan and Smith, Noah A and Ma, Wei-Chiu and Krishna, Ranjay},
  journal={arXiv preprint arXiv:2404.12390},
  year={2024}
}

@misc{lyu2025jigsawpuzzles,
      title={Jigsaw-Puzzles: From Seeing to Understanding to Reasoning in Vision-Language Models}, 
      author={Zesen Lyu and Dandan Zhang and Wei Ye and Fangdi Li and Zhihang Jiang and Yao Yang},
      year={2025},
      eprint={2505.20728},
      archivePrefix={arXiv},
      primaryClass={cs.AI},
      url={https://arxiv.org/abs/2505.20728}, 
}

@article{ren2025vgrp,
  title={{VGRP-Bench}: Visual Grid Reasoning Puzzle Benchmark for Large Vision-Language Models},
  author={Ren, Yufan and Tertikas, Konstantinos and Maiti, Shalini and Han, Junlin and Zhang, Tong and S{\"u}sstrunk, Sabine and Kokkinos, Filippos},
  journal={arXiv preprint arXiv:2503.23064},
  year={2025}
}

@article{tang2025lego,
  title={{LEGO-Puzzles}: How Good Are MLLMs at Multi-Step Spatial Reasoning?},
  author={Tang, Kexian and Gao, Junyao and Zeng, Yanhong and Duan, Haodong and Sun, Yanan and Xing, Zhening and Liu, Wenran and Lyu, Kaifeng and Chen, Kai},
  journal={arXiv preprint arXiv:2503.19990},
  year={2025}
}

@article{xu2025vlhsa,
  title={{VLHSA}: Vision-Language Hierarchical Semantic Alignment for Jigsaw Puzzle Solving with Eroded Gaps},
  author={Xu, Zhuoning and Liu, Xinyan},
  journal={arXiv preprint arXiv:2509.25202},
  year={2025}
}

@article{elkin2025puzlm,
  title={{PuzLM}: Solving Jigsaw Puzzles with Sequence-to-Sequence Language Models},
  author={Elkin, Gur and Shahar, Ofir Itzhak and Ben-Shahar, Ohad},
  journal={arXiv preprint arXiv:2511.06315},
  year={2025}
}

@inproceedings{scarpellini2024diffassemble,
  title={{DiffAssemble}: A Unified Graph-Diffusion Model for 2D and 3D Reassembly},
  author={Scarpellini, Gianluca and Fiorini, Stefano and Giuliari, Francesco and Morerio, Pietro and Del Bue, Alessio},
  booktitle={CVPR},
  year={2024}
}

@inproceedings{agustsson2017div2k,
  title={{NTIRE} 2017 Challenge on Single Image Super-Resolution: Dataset and Study},
  author={Agustsson, Eirikur and Timofte, Radu},
  booktitle={CVPR Workshops},
  year={2017}
}

@inproceedings{gu2019div8k,
  title={{DIV8K}: DIVerse 8K Resolution Image Dataset},
  author={Gu, Shuhang and Lugmayr, Andreas and Danelljan, Martin and Fritsche, Manuel and Lamour, Julien and Timofte, Radu},
  booktitle={ICCV Workshops},
  year={2019}
}

@misc{unsplash,
  title={Unsplash},
  author={{Unsplash Inc.}},
  howpublished={\url{https://unsplash.com}},
  year={2024},
  note={High-resolution photographs under Unsplash License}
}

@article{lanczos,
  title={Applied Analysis},
  author={Lanczos, Cornelius},
  journal={Prentice-Hall},
  year={1956}
}

@article{bai2025qwen25vl,
  title={{Qwen2.5-VL} Technical Report},
  author={Bai, Shuai and Chen, Keqin and Liu, Xuejing and Wang, Jialin and Ge, Wenbin and Song, Sibo and Dang, Kai and Wang, Peng and Wang, Shijie and Tang, Jun and others},
  journal={arXiv preprint arXiv:2502.13923},
  year={2025}
}

@article{team2024gemma,
  title={Gemma 2: Improving Open Language Models at a Practical Size},
  author={{Google DeepMind}},
  journal={arXiv preprint arXiv:2408.00118},
  year={2024}
}

@article{liu2023llava,
  title={{Visual Instruction Tuning}},
  author={Liu, Haotian and Li, Chunyuan and Wu, Qingyang and Lee, Yong Jae},
  journal={NeurIPS},
  year={2023}
}

@article{dai2023instructblip,
  title={{InstructBLIP}: Towards General-purpose Vision-Language Models with Instruction Tuning},
  author={Dai, Wenliang and Li, Junnan and Li, Dongxu and Tiong, Anthony Meng Huat and Zhao, Junqi and Wang, Weisheng and Li, Boyang and Fung, Pascale and Hoi, Steven},
  journal={NeurIPS},
  year={2023}
}

@article{li2023blip2,
  title={{BLIP-2}: Bootstrapping Language-Image Pre-training with Frozen Image Encoders and Large Language Models},
  author={Li, Junnan and Li, Dongxu and Savarese, Silvio and Hoi, Steven},
  journal={ICML},
  year={2023}
}

@article{pomeranz2011fully,
  title={A Fully Automated Greedy Square Jigsaw Puzzle Solver},
  author={Pomeranz, Dolev and Shemesh, Michal and Ben-Shahar, Ohad},
  journal={CVPR},
  year={2011}
}

@article{sholomon2013genetic,
  title={A Genetic Algorithm-Based Solver for Very Large Jigsaw Puzzles},
  author={Sholomon, Dror and David, Omid and Netanyahu, Nathan S},
  journal={CVPR},
  year={2013}
}

@article{son2014solving,
  title={Solving Square Jigsaw Puzzles with Loop Constraints},
  author={Son, Kilho and Hays, James and Cooper, David B},
  journal={ECCV},
  year={2014}
}

@article{johnson2017clevr,
  title={{CLEVR}: A Diagnostic Dataset for Compositional Language and Elementary Visual Reasoning},
  author={Johnson, Justin and Hariharan, Bharath and van der Maaten, Laurens and Fei-Fei, Li and Zitnick, C Lawrence and Girshick, Ross},
  journal={CVPR},
  year={2017}
}

@article{hudson2019gqa,
  title={{GQA}: A New Dataset for Real-World Visual Reasoning and Compositional Question Answering},
  author={Hudson, Drew A and Manning, Christopher D},
  journal={CVPR},
  year={2019}
}

@article{dosovitskiy2021vit,
  title={An Image is Worth 16x16 Words: Transformers for Image Recognition at Scale},
  author={Dosovitskiy, Alexey and Beyer, Lucas and Kolesnikov, Alexander and Weissenborn, Dirk and Zhai, Xiaohua and Unterthiner, Thomas and Dehghani, Mostafa and Minderer, Matthias and Heigold, Georg and Gelly, Sylvain and others},
  journal={ICLR},
  year={2021}
}

@article{radford2021clip,
  title={Learning Transferable Visual Models From Natural Language Supervision},
  author={Radford, Alec and Kim, Jong Wook and Hallacy, Chris and Ramesh, Aditya and Goh, Gabriel and Agarwal, Sandhini and Sastry, Girish and Askell, Amanda and Mishkin, Pamela and Clark, Jack and others},
  journal={ICML},
  year={2021}
}

@article{thrush2022winoground,
  title={{Winoground}: Probing Vision and Language Models for Visio-Linguistic Compositionality},
  author={Thrush, Tristan and Jiang, Ryan and Bartolo, Max and Singh, Amanpreet and Williams, Adina and Kiela, Douwe and Ross, Candace},
  journal={CVPR},
  year={2022}
}

@inproceedings{parcalabescu2022valse,
    title = "{VALSE}: A Task-Independent Benchmark for Vision and Language Models Centered on Linguistic Phenomena",
    author = "Parcalabescu, Letitia  and
      Cafagna, Michele  and
      Muradjan, Lilitta  and
      Frank, Anette  and
      Calixto, Iacer  and
      Gatt, Albert",
    booktitle = "ACL",
    month = may,
    year = "2022",
    pages = "8253--8280",
}

@inproceedings{noroozi2016unsupervised,
  title={Unsupervised Learning of Visual Representations by Solving Jigsaw Puzzles},
  author={Noroozi, Mehdi and Favaro, Paolo},
  booktitle={European Conference on Computer Vision (ECCV)},
  pages={69--84},
  year={2016}
}

@inproceedings{ren2023masked,
  title={Masked Jigsaw Puzzle: A Versatile Position Embedding for Vision Transformers},
  author={Ren, Bin and Liu, Yahui and Song, Yue and Bi, Wei and Cucchiara, Rita and Sebe, Nicu and Wang, Wei},
  booktitle={IEEE/CVF Conference on Computer Vision and Pattern Recognition (CVPR)},
  pages={20382--20391},
  year={2023}
}

@inproceedings{carlucci2019domain,
  title={Domain Generalization by Solving Jigsaw Puzzles},
  author={Carlucci, Fabio Maria and D'Innocente, Antonio and Bucci, Silvia and Caputo, Barbara and Tommasi, Tatiana},
  booktitle={IEEE/CVF Conference on Computer Vision and Pattern Recognition (CVPR)},
  pages={2229--2238},
  year={2019}
}

@article{wang2025jigsawr1,
  title={Jigsaw-R1: A Study of Rule-based Visual Reinforcement Learning with Jigsaw Puzzles},
  author={Wang, Zifu and Zhu, Junyi and Tang, Bo and Li, Zhiyu and Xiong, Feiyu and Yu, Jiaqian and Blaschko, Matthew B.},
  journal={Transactions on Machine Learning Research},
  year={2025}
}

@article{jeddi2025puzzlecraft,
  title={PuzzleCraft: Exploration-Aware Curriculum Learning for Puzzle-Based {RLVR} in {VLMs}},
  author={Jeddi, Ahmadreza and Karaimer, Hakki Can and Nguyen, Hue and Wang, Zhongling and Zhao, Ke and Rajabi, Javad and Zhang, Ran and Goyal, Raghav and Derpanis, Konstantinos G. and Taati, Babak and Grzeszczuk, Radek},
  journal={arXiv preprint arXiv:2512.14944},
  year={2025}
}

@article{liu2025spatialssrl,
  title={Spatial-{SSRL}: Enhancing Spatial Understanding via Self-Supervised Reinforcement Learning},
  author={Liu, Yuhong and Zhang, Beichen and Zang, Yuhang and Cao, Yuhang and Xing, Long and Dong, Xiaoyi and Duan, Haodong and Lin, Dahua and Wang, Jiaqi},
  journal={arXiv preprint arXiv:2510.27606},
  year={2025}
}

@article{song2023sd2rl, 
title={Siamese-Discriminant Deep Reinforcement Learning for Solving Jigsaw Puzzles with Large Eroded Gaps}, 
volume={37}, 
number={2}, 
journal={AAAI}, 
author={Song, Xingke and Jin, Jiahuan and Yao, Chenglin and Wang, Shihe and Ren, Jianfeng and Bai, Ruibin}, 
year={2023}, 
month={Jun.}, 
pages={2303–2311} 
}

@article{Li_Ji_Wu_Li_Qin_Wei_Zimmermann_2024, 
title={Panoptic Scene Graph Generation with Semantics-Prototype Learning}, 
volume={38},
DOI={10.1609/aaai.v38i4.28098}, 
number={4}, 
journal={AAAI}, 
author={Li, Li and Ji, Wei and Wu, Yiming and Li, Mengze and Qin, You and Wei, Lina and Zimmermann, Roger}, 
year={2024}, 
month={Mar.}, 
pages={3145-3153} }

@InProceedings{Li_2025_CVPR,
    author    = {Li, Shawn and Gong, Huixian and Dong, Hao and Yang, Tiankai and Tu, Zhengzhong and Zhao, Yue},
    title     = {DPU: Dynamic Prototype Updating for Multimodal Out-of-Distribution Detection},
    booktitle = {CVPR},
    month     = {June},
    year      = {2025},
    pages     = {10193-10202}
}

@InProceedings{li2025secureondevicevideoood,
    title={Secure On-Device Video OOD Detection Without Backpropagation}, 
    author={Shawn Li and Peilin Cai and Yuxiao Zhou and Zhiyu Ni and Renjie Liang and You Qin and Yi Nian and Zhengzhong Tu and Xiyang Hu and Yue Zhao},
    booktitle = {ICCV},
    month     = {October},
    year      = {2025}
}

@inproceedings{li-etal-2025-treble,
    title = "Treble Counterfactual {VLM}s: A Causal Approach to Hallucination",
    author = "Shawn, Li  and
      Qu, Jiashu  and
      Song, Linxin  and
      Zhou, Yuxiao  and
      Qin, Yuehan  and
      Yang, Tiankai  and
      Zhao, Yue",
    booktitle = "EMNLP",
    month = nov,
    year = "2025",
}

@inproceedings{li2026defensespromptattackslearn,
    title={Defenses Against Prompt Attacks Learn Surface Heuristics}, 
    author={Shawn Li and Chenxiao Yu and Zhiyu Ni and Hao Li and Charith Peris and Chaowei Xiao and Yue Zhao},
    year={2026},
    booktitle = "ACL",
    year = "2026",

}

@misc{li2026geometrydensityfewshotcrossdomain,
      title={Geometry over Density: Few-Shot Cross-Domain OOD Detection}, 
      author={Shawn Li and You Qin and Jiate Li and Charith Peris and Lisa Bauer and Roger Zimmermann and Yue Zhao},
      year={2026},
      eprint={2605.03410},
      archivePrefix={arXiv},
      primaryClass={cs.AI},
      url={https://arxiv.org/abs/2605.03410}, 
}

@misc{li2026fortisbenchmarkingoverprivilegeagent,
      title={FORTIS: Benchmarking Over-Privilege in Agent Skills}, 
      author={Shawn Li and Chenxiao Yu and Han Wang and Wei Yang and Ryan Rossi and Franck Dernoncourt and Xiyang Hu and Philip Yu and Chaowei Xiao and Huan Zhang and Yue Zhao},
      year={2026},
      eprint={2605.09163},
      archivePrefix={arXiv},
      primaryClass={cs.AI},
      url={https://arxiv.org/abs/2605.09163}, 
}
\bibliographystyle{iclr2026_conference}

\clearpage
\appendix

\section{Generation Pipeline Details}
\label{app:pipeline}

\paragraph{Shape version.}
Given source image $I$ and grid size $N$, we resize $I$ so dimensions are divisible by $N$. For each internal edge, we randomly sample polarity (tab-blank or blank-tab); boundary edges are flat. Tabs are semicircular with radius $r = \text{tab\_ratio} \times \min(h, w) / 2$. Content is extracted with tab regions carrying neighboring pixels. Piece IDs (0 to $N^2{-}1$) are randomly permuted, and pieces are arranged by ID with numeric labels. The pipeline is deterministic given a seed.

\paragraph{No-shape version (for ablation).}
For the ablation study, we generate a \textbf{no-shape} (rectangular) version of the 4$\times$4 test set. The no-shape pipeline uses the same source images, the same random seed (thus identical piece ID permutations), and the same grid layout, but replaces tab-and-blank edges with straight cuts. This produces rectangular pieces without geometric interlocking. The \emph{only} difference between shape and no-shape versions is the presence of tab-blank boundaries, enabling controlled measurement of whether models utilize geometric constraints. We restrict no-shape generation to 4$\times$4 because larger grids with rectangular cuts have ambiguous ground truth: visually similar regions (e.g., sky, grass) become interchangeable, making multiple arrangements equally valid.

\section{Prompt Templates}
\label{app:prompts}

All models receive identical prompts consisting of a system prompt and a user prompt. The system prompt establishes the task context and emphasizes that piece IDs are randomly assigned. The user prompt specifies the grid size and output format. We provide separate prompts for shape-constrained and no-shape (ablation) conditions.

\paragraph{Shape-Constrained Prompts.}

\noindent\textbf{System Prompt:}
\begin{lstlisting}[basicstyle=\ttfamily\scriptsize,breaklines=true,breakatwhitespace=true]
You are an expert visual puzzle solver. You must carefully analyze the image provided and determine the correct arrangement of puzzle pieces based on:
1. The visual content of each piece (colors, textures, patterns, objects)
2. The shape of piece edges (tabs and blanks must match between adjacent pieces)
3. Corner pieces have 2 flat edges, edge pieces have 1 flat edge

CRITICAL: The pieces are RANDOMLY SHUFFLED with RANDOM labels. The piece labeled "0" is NOT necessarily at position [0,0]. The piece labeled "3" is NOT necessarily a corner piece. You MUST look at the ACTUAL IMAGE of each piece to determine its correct position. Do NOT assume any relationship between piece numbers and positions.
\end{lstlisting}

\noindent\textbf{User Prompt:}
\begin{lstlisting}[basicstyle=\ttfamily\scriptsize,breaklines=true,breakatwhitespace=true]
This is a NxN jigsaw puzzle with N^2 pieces labeled 0 to N^2-1.

Analyze the image carefully:
1. Identify the 4 corner pieces (2 flat edges each)
2. Identify the edge pieces (1 flat edge each)
3. Match pieces by their visual content and edge shapes

Output ONLY a JSON object mapping piece_id to [row, col]:
{"0": [row, col], "1": [row, col], ...}

Rows and columns are 0-indexed from top-left.
\end{lstlisting}

\paragraph{No-Shape Prompts (Ablation).}

\noindent\textbf{System Prompt:}
\begin{lstlisting}[basicstyle=\ttfamily\scriptsize,breaklines=true,breakatwhitespace=true]
You are an expert visual puzzle solver. You must carefully analyze the image provided and determine the correct arrangement of puzzle pieces based on:
1. The visual content of each piece (colors, textures, patterns, objects)
2. How adjacent pieces connect visually (matching edges, continuous patterns, color continuity)

NOTE: These puzzle pieces are simple rectangles without interlocking shapes. You must rely ONLY on visual content to determine the correct arrangement.

CRITICAL: The pieces are RANDOMLY SHUFFLED with RANDOM labels. The piece labeled "0" is NOT necessarily at position [0,0]. You MUST look at the ACTUAL IMAGE of each piece to determine its correct position. Do NOT assume any relationship between piece numbers and positions.
\end{lstlisting}

\noindent\textbf{User Prompt:}
\begin{lstlisting}[basicstyle=\ttfamily\scriptsize,breaklines=true,breakatwhitespace=true]
This is a NxN jigsaw puzzle with N^2 rectangular pieces labeled 0 to N^2-1.

Analyze the image carefully:
1. Look at the visual content of each piece (colors, textures, patterns, objects)
2. Determine how pieces connect by matching visual content at edges
3. Reconstruct the original image by finding the correct position for each piece

Output ONLY a JSON object mapping piece_id to [row, col]:
{"0": [row, col], "1": [row, col], ...}

Rows and columns are 0-indexed from top-left.
\end{lstlisting}

\section{Experiment Details}
\label{app:experiment}

\paragraph{Frontier VLMs (Zero-Shot).}
We evaluate five frontier VLMs in six zero-shot configurations with identical prompts: GPT-5.5 and GPT-5.4-mini (OpenAI, May 2025); Grok-4.2 with and without extended reasoning (xAI); Llama-4-Maverick-17B-128E-Instruct-FP8 (Meta); Claude Opus 4.8 (Anthropic). All models support high-resolution image input and 128K+ token context windows, sufficient for processing 16$\times$16 puzzle layouts with 256 labeled pieces. For models with reasoning capabilities, we use their reasoning-enabled modes: GPT-5.5 performs internal chain-of-thought reasoning before outputting JSON (similar to o1/o3 models), and Grok-4.2-reasoning uses \texttt{reasoning\_effort: medium}. While our prompt requests JSON-only output for parsing, these models perform extended reasoning internally. We separately evaluate Grok without reasoning mode as a control, showing only modest improvement from explicit reasoning (11.97\% vs.\ 6.90\% PA on 4$\times$4).

\paragraph{Fine-Tuned Models.}
We fine-tune Qwen3-VL-8B and Gemma3-12B on a mixed-grid training set with weighted sampling (4$\times$4: 3K, 8$\times$8: 2K, 12$\times$12: 1K, 16$\times$16: 0.5K; 6.5K total). Training uses 1 epoch, learning rate $1\times10^{-5}$, effective batch size 64, BF16 mixed precision, and Flash Attention 2.

\paragraph{Evaluation Protocol.}
All models are evaluated on the eval set (250 instances per grid size, 1,000 total). Model outputs are parsed to extract piece ID to position mappings; responses with invalid JSON receive zero accuracy on all metrics.

\section{Evaluation Metrics}
\label{app:metrics}

We define four evaluation metrics to comprehensively assess model performance on jigsaw puzzle solving. Let $N$ denote the grid size, $\hat{p}: \{0, \ldots, N^2-1\} \rightarrow \{0, \ldots, N-1\}^2$ denote the predicted position mapping, and $p^*$ denote the ground-truth mapping.

\paragraph{Piece Accuracy (PA).}
The fraction of pieces placed in their correct positions:
\begin{equation}
\text{PA} = \frac{1}{N^2} \sum_{i=0}^{N^2-1} \mathbf{1}[\hat{p}(i) = p^*(i)]
\end{equation}
PA measures per-piece correctness without considering spatial relationships between pieces.

\paragraph{Exact Match (EM).}
A binary indicator of whether the entire puzzle is solved correctly:
\begin{equation}
\text{EM} = \mathbf{1}[\forall i: \hat{p}(i) = p^*(i)]
\end{equation}
EM equals 1 only when all $N^2$ pieces are in their correct positions (PA = 100\%).

\paragraph{Adjacency Accuracy (AA).}
The fraction of correctly predicted adjacent piece pairs. We define the ground-truth adjacency set $\mathcal{A}^*$ as all piece ID pairs $(i, j)$ where $i$ and $j$ are horizontally or vertically adjacent in the solved puzzle. For an $N \times N$ grid, $|\mathcal{A}^*| = 2N(N-1)$. The predicted adjacency set $\hat{\mathcal{A}}$ is computed analogously from $\hat{p}$:
\begin{equation}
\text{AA} = \frac{|\hat{\mathcal{A}} \cap \mathcal{A}^*|}{|\mathcal{A}^*|}
\end{equation}
AA measures local coherence: a model can achieve high AA by correctly grouping pieces locally even if global positions are wrong. Under a uniformly random permutation, the probability that any given ground-truth adjacent pair $(i,j)$ lands on an adjacent position pair equals the number of adjacent position pairs divided by the total number of ordered position pairs: $\mathbb{E}[\text{AA}] = \frac{2 \cdot 2N(N-1)}{N^2(N^2-1)} = \frac{4}{N(N+1)}$, yielding baselines of 20.0\% (4$\times$4), 5.6\% (8$\times$8), 2.6\% (12$\times$12), and 1.5\% (16$\times$16).

\paragraph{Shape Compatibility (SC).}
The fraction of adjacent piece placements with geometrically compatible edges. Each piece carries edge types from its \emph{original} grid position: internal edges are randomly assigned tab or blank, while boundary edges are flat (\S\ref{sec:shape}). Let $\mathcal{E}^{\hat{p}}$ be the set of internal edges in the predicted arrangement, and $c(e) = 1$ if the two facing edges are complementary (tab--blank):
\begin{equation}
\text{SC} = \frac{1}{|\mathcal{E}^{\hat{p}}|} \sum_{e \in \mathcal{E}^{\hat{p}}} c(e)
\end{equation}
Under random placement, SC $\neq$ 50\% because boundary pieces carry flat edges that are always incompatible when placed at internal positions. Consider a random internal grid edge where piece $i$ contributes its right edge and piece $j$ contributes its left edge. Among $N^2$ pieces, $N(N{-}1)/2$ have tab-right and $N(N{-}1)/2$ have blank-left (the counts are equal because each horizontal edge assignment creates one of each). Let $T_R$ denote the number of tab-right pieces; since tab-right and blank-left counts coincide ($T_R = B_L$), sampling two distinct pieces without replacement gives $\mathbb{E}[\text{SC}] = 2\,\mathbb{E}[T_R(T_R{-}X)]/(N^2(N^2{-}1))$, where $X$ counts pieces that are \emph{both} tab-right and blank-left. Evaluating $\mathbb{E}[T_R^2]$ via $\text{Var}[T_R]{+}(\mathbb{E}[T_R])^2$ and $\mathbb{E}[X] = N(N{-}2)/4$ yields baselines of 30.8\% (4$\times$4), 39.0\% (8$\times$8), 42.3\% (12$\times$12), and 44.1\% (16$\times$16), confirmed by Monte Carlo simulation ($2{\times}10^6$ trials). The simple approximation $\frac{1}{2}((N{-}1)/N)^2$ underestimates by ignoring the without-replacement correction. SC above the grid-specific baseline indicates the model utilizes shape constraints.

\section{Additional Results}
\label{app:results}

\subsection{Weighted SFT Training}

We explore whether emphasizing harder training samples during fine-tuning can improve performance on larger grid sizes. We implement an inverse-accuracy weighting scheme where training samples are weighted by $(1 - \text{PA}_{\text{grid}})$, upweighting harder grids (12$\times$12, 16$\times$16) while downweighting easier grids (4$\times$4) where baseline performance is already high. We train Qwen3-VL-8B for 3 epochs on a mixed-grid training set (20K samples) with these sample weights applied to the loss function.

As shown in Tab.~\ref{tab:weighted_sft}, the weighted approach performs significantly worse across all grid sizes and all metrics. These results suggest that the scaling cliff observed in Sec.~5 reflects a fundamental \textbf{architectural limitation} rather than a data distribution issue. Reweighting training samples to emphasize harder grids does not help (and in fact hurts) because the underlying model architecture cannot learn to solve larger puzzles regardless of how training signal is distributed. This finding reinforces our conclusion that current VLM architectures lack the compositional reasoning capabilities required for scalable geometric constraint satisfaction, and that progress on larger grids will likely require architectural innovations rather than training data scaling strategies.

\begin{table}[t]
\centering
\caption{Comparison of baseline SFT vs.\ weighted SFT on Qwen3-VL-8B across all metrics. PA: Piece Accuracy (\%), EM: Exact Match (\%), AA: Adjacency Accuracy (\%), SC: Shape Compatibility (\%). Weighted SFT uses inverse-accuracy sample weighting to emphasize harder grids but performs worse across all settings.}
\label{tab:weighted_sft}
\small
\begin{tabular}{l|cccc|cccc}
\toprule
& \multicolumn{4}{c|}{\textbf{Baseline SFT}} & \multicolumn{4}{c}{\textbf{Weighted SFT}} \\
\textbf{Grid} & PA & EM & AA & SC & PA & EM & AA & SC \\
\midrule
4$\times$4 & 97.28 & 90.80 & 95.77 & 98.40 & 56.40 & 0.80 & 31.68 & 67.90 \\
8$\times$8 & 27.34 & 0.00 & 15.58 & 70.75 & 9.12 & 0.00 & 1.09 & 48.39 \\
12$\times$12 & 3.44 & 0.00 & 0.59 & 52.11 & 0.89 & 0.00 & 0.00 & 50.00 \\
16$\times$16 & 0.40 & 0.00 & 0.00 & 45.02 & 0.00 & 0.00 & 0.00 & 50.00 \\
\bottomrule
\end{tabular}
\end{table}

\subsection{Classical Solver Baseline}
\label{app:classical_solver}

To verify that every puzzle instance has a unique solution, we implement a classical constraint-based solver that uses exact edge signature information. Each piece has four edge types (tab, blank, or flat). The solver operates as follows:

\paragraph{Edge Compatibility Rules.}
Adjacent pieces must have complementary edges: a \emph{tab} fits into a \emph{blank}, while \emph{flat} edges occur only at puzzle boundaries (corners have two flat edges, border pieces have one).

\paragraph{Algorithm.}
We use backtracking search with constraint propagation. First, we precompute candidate pieces for each grid position based on boundary constraints (e.g., corner positions require exactly two flat edges). Then we fill positions row by row, checking at each step that the placed piece is compatible with all already-placed neighbors. If no valid piece exists, we backtrack.

\paragraph{Results.}
The solver achieves 100\% accuracy across all grid sizes (4$\times$4, 8$\times$8, 12$\times$12, 16$\times$16) on all 500 test instances per size, confirming that every puzzle in \ours{} has a unique solution.

\subsection{Auxiliary Hints Do Not Help}

We investigate whether providing auxiliary information can help models solve larger puzzles. We design two types of hints: (1) partial ground truth positions, and (2) visual similarity groupings.

\paragraph{Partial Ground Truth (Clue Benchmark).}
We test whether models can leverage partial solutions by revealing some pieces' correct positions upfront. Tab.~\ref{tab:clue_benchmark} shows results on 8$\times$8 puzzles with varying amounts of known information.

\begin{table}[t]
\centering
\caption{Effect of visual similarity hints on 8$\times$8 puzzles (Qwen SFT).}
\label{tab:seg_hint}
\small
\begin{tabular}{lc}
\toprule
\textbf{Condition} & \textbf{PA} \\
\midrule
Without hint & 27.34\% \\
With hint & 18.54\% \\
\midrule
Difference & $-$10.52\% \\
\bottomrule
\end{tabular}
\end{table}

\begin{table}[t]
\centering
\caption{Clue benchmark results on 8$\times$8 puzzles. We provide models with $n$ pieces already at correct positions and measure PA only on the remaining puzzle.}
\label{tab:clue_benchmark}
\small
\begin{tabular}{lccccc}
\toprule
\textbf{Clue Type} & \textbf{Known} & \textbf{Unknown} & \textbf{Qwen SFT} & \textbf{GPT-5.5} \\
\midrule
corners & 4 & 60 & 26.74\% & 2.31\% \\
edges & 28 & 36 & 18.76\% & 3.76\% \\
diagonal & 16 & 48 & 22\% & 2.02\% \\
two\_rows & 16 & 48 & 23\% & 2.62\% \\
center4x4 & 16 & 48 & 21\% & 3.02\% \\
extreme60 & 60 & 4 & 25.0\% & 2.50\% \\
\bottomrule
\end{tabular}
\end{table}

The extreme60 setting is particularly revealing: even when 60 out of 64 pieces are already correctly placed and the model only needs to determine 4 remaining pieces, GPT-5.5 achieves only 25\% PA. Further analysis shows the model places these 4 pieces within the correct 2$\times$2 region but swaps their exact positions. This confirms models can read and understand hint information but struggle with precise spatial reasoning.

\begin{figure}[h]
\centering
\includegraphics[width=0.95\linewidth]{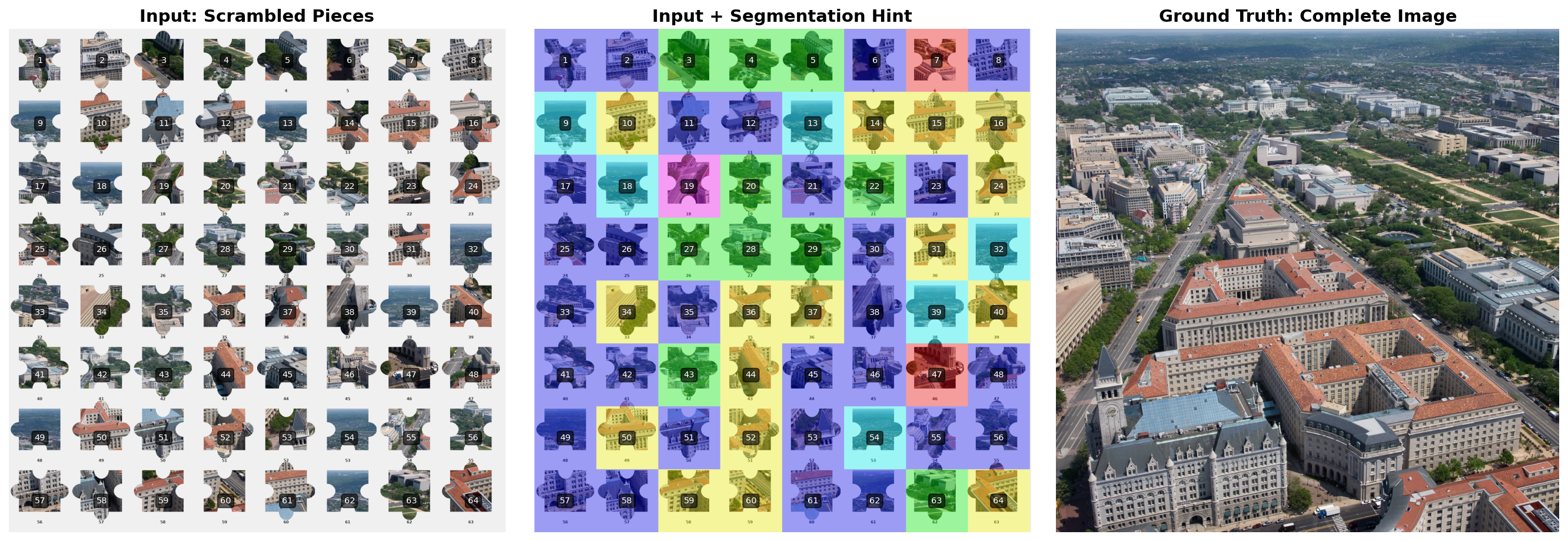}
\caption{Visual similarity hint. Left: original scrambled pieces. Middle: pieces colored by cluster membership (same color = visually similar). Right: ground truth.}
\label{fig:seg_hint}
\end{figure}

\paragraph{Visual Similarity Hints.}
Standard segmentation maps have limited applicability to jigsaw puzzles since we need piece-level grouping rather than pixel-level segmentation. We design an alternative approach: clustering pieces by visual similarity (color histograms in LAB space) and conveying this grouping through natural language to avoid adding visual processing burden. Fig.~\ref{fig:seg_hint} illustrates the hint visualization.

\begin{figure*}[h]
\centering
\begin{tabular}{cc}
\includegraphics[width=0.45\linewidth]{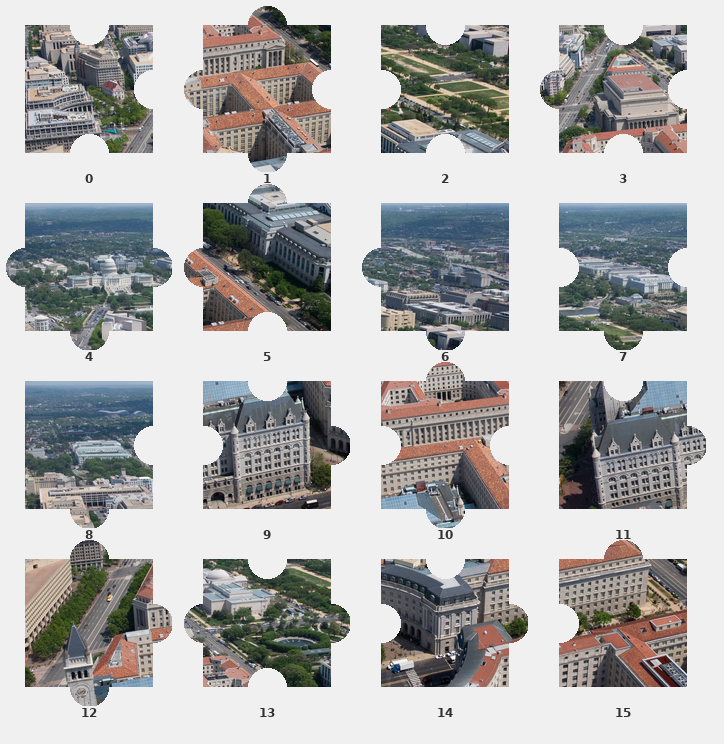} &
\includegraphics[width=0.45\linewidth]{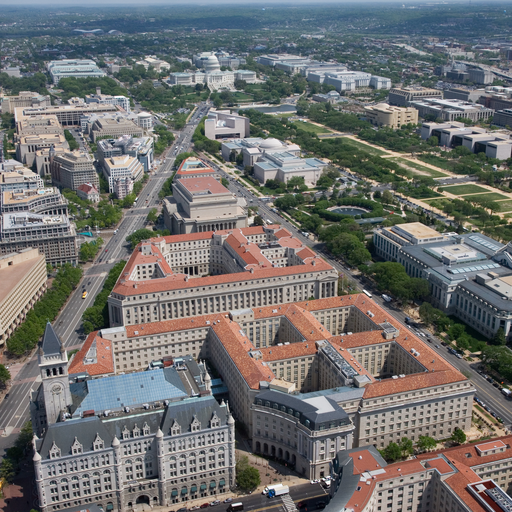} \\
\multicolumn{2}{c}{(a) 4$\times$4 grid: 16 pieces} \\[1em]
\includegraphics[width=0.45\linewidth]{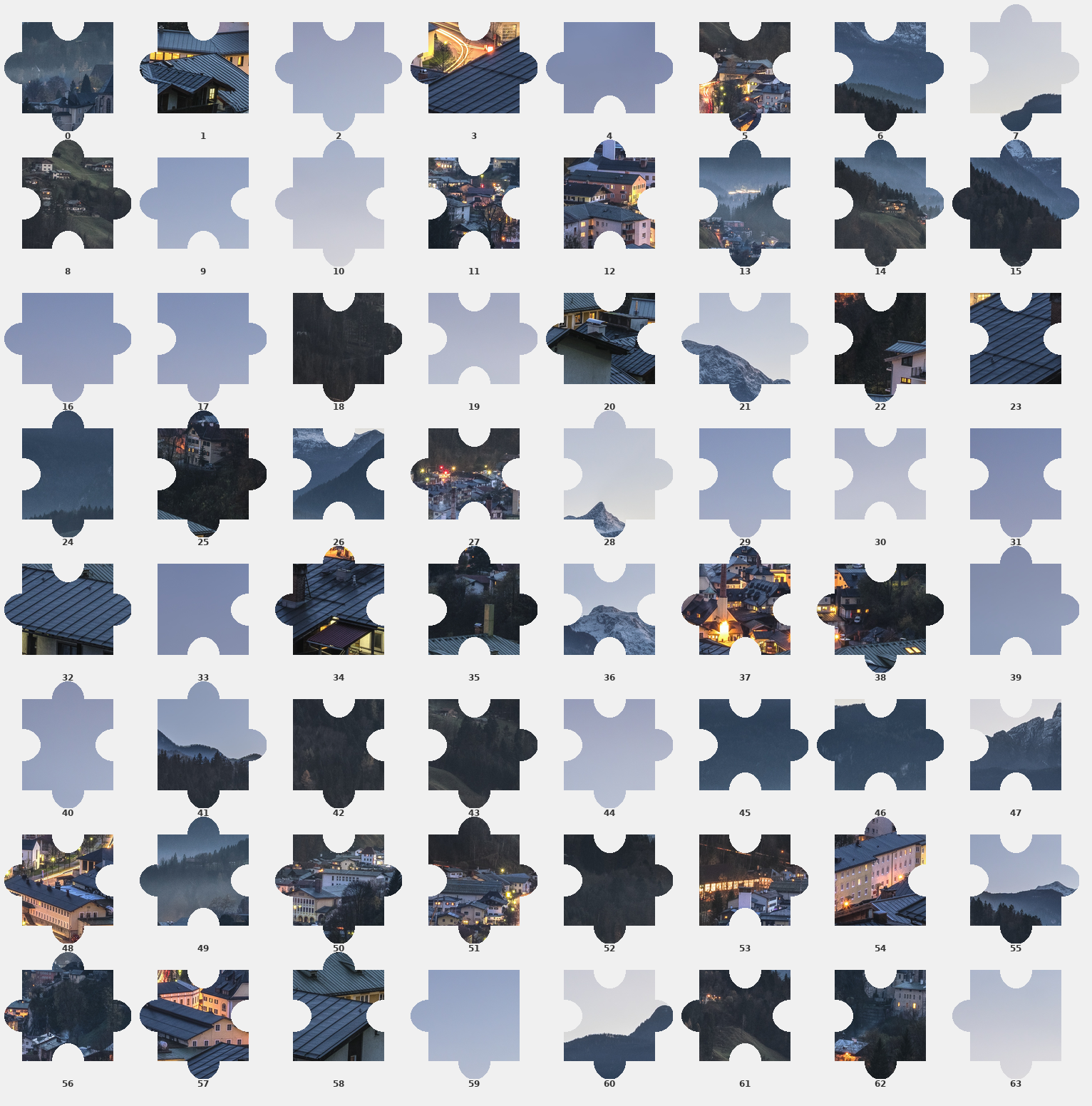} &
\includegraphics[width=0.45\linewidth]{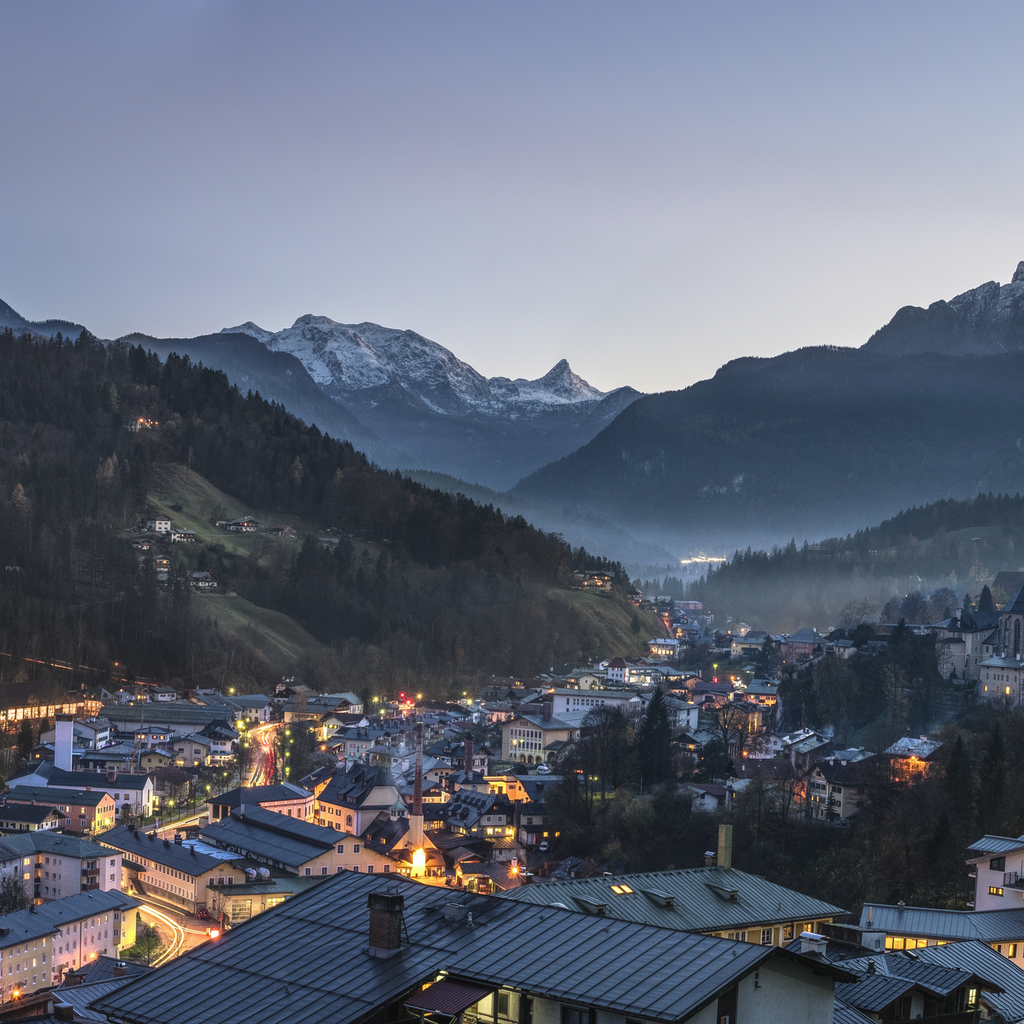} \\
\multicolumn{2}{c}{(b) 8$\times$8 grid: 64 pieces} \\[1em]
\includegraphics[width=0.45\linewidth]{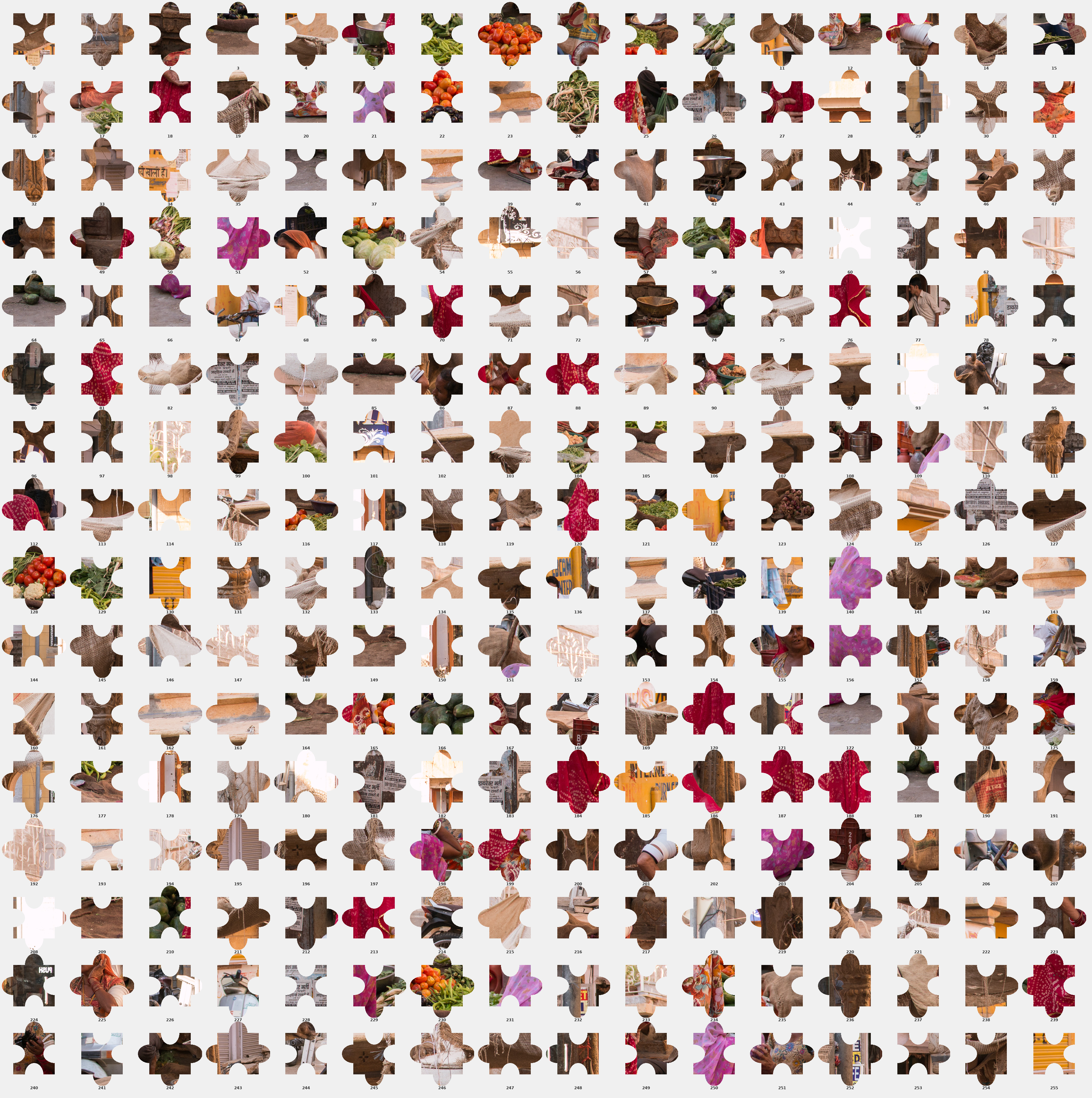} &
\includegraphics[width=0.45\linewidth]{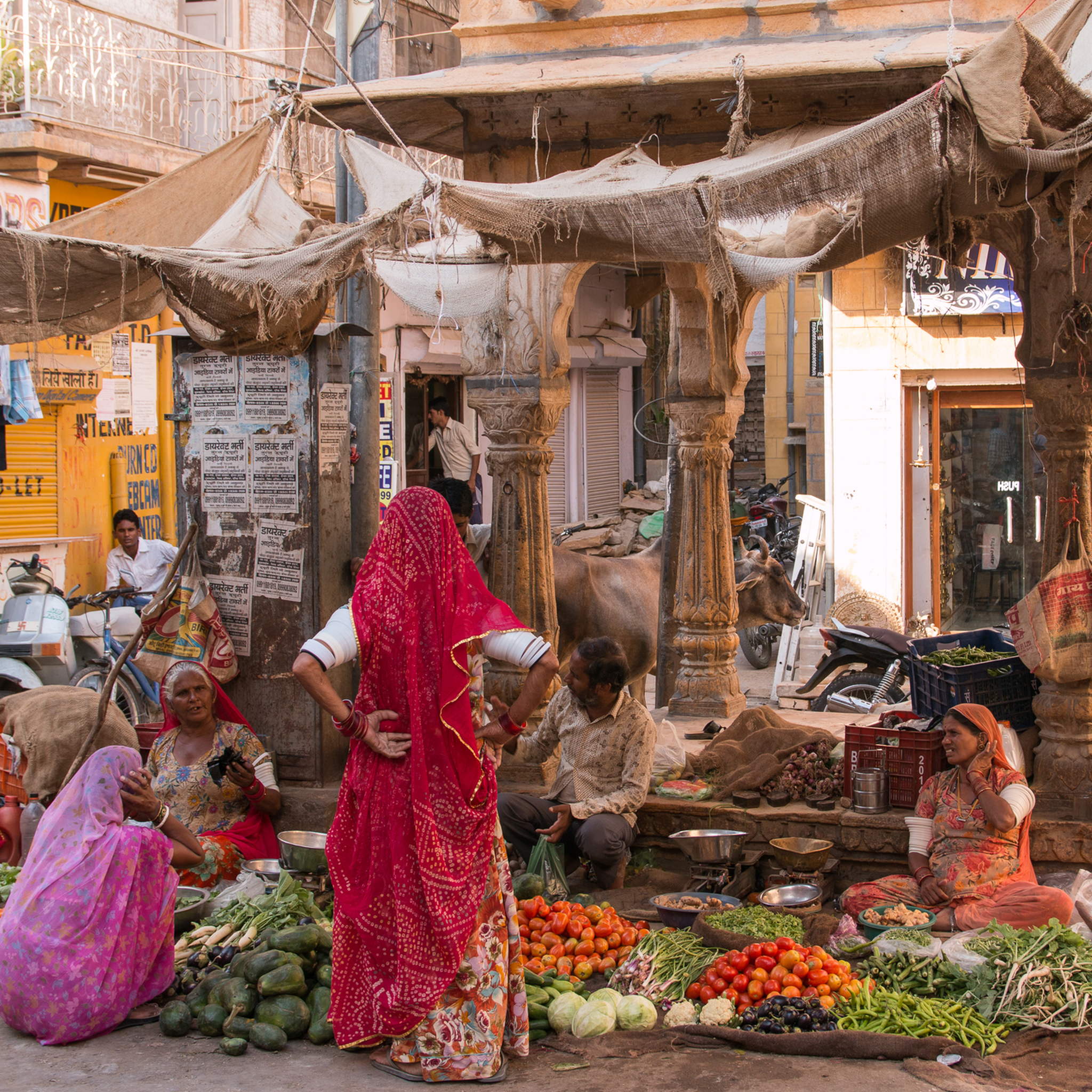} \\
\multicolumn{2}{c}{(c) 16$\times$16 grid: 256 pieces} \\
\end{tabular}
\caption{Example instances from \ours{}. Left: shuffled puzzle layout with labeled pieces. Right: original source image. As grid size increases, the number of pieces grows quadratically, making constraint satisfaction exponentially harder. At 16$\times$16, models must reason about 256 pieces with 480 internal edges.}
\label{fig:examples_appendix}
\end{figure*}

As shown in Tab.~\ref{tab:seg_hint}, the visual similarity hint actually \emph{hurts} performance by over 10\%. We hypothesize two reasons for this failure. First, VLM pretraining data rarely contains such visually dense, spatially structured images. A single 8$\times$8 puzzle packs 64 distinct regions with complex geometric relationships, far outside typical training distributions. Second, the image already encodes rich visual content plus geometric cues (tabs, blanks, edge shapes). Adding textual hints creates information overload: the model must simultaneously process dense visual features, geometric constraints, and auxiliary guidance, exceeding its effective integration capacity.

These experiments suggest the bottleneck is not hint quality but a fundamental limitation in how current VLMs process spatially dense, geometrically complex inputs. Providing additional hints, whether oracle positions or similarity groupings, cannot compensate for this architectural gap.

\section{Dataset Examples}
\label{app:examples}

We present example instances from \ours{} at different grid sizes. Each row shows the shuffled puzzle layout (left) that serves as input to the model, and the original source image (right) that represents the ground-truth solution. Piece IDs are randomly assigned and displayed on each piece. The model must determine the correct position mapping from piece ID to grid location. See Fig.~\ref{fig:example_main} in the main text for a 12$\times$12 example.

\section{Dataset Licensing}
\label{app:license}

\textbf{DIV2K/DIV8K} are released by ETH Zürich under academic license for research purposes. \textbf{Unsplash} images are licensed under the Unsplash License, permitting free use.

\end{document}